\documentclass{article} % For LaTeX2e
\usepackage{iclr2027_conference,times}

\usepackage{amsmath,amsfonts,bm}

\def\eqref#1{equation~\ref{#1}}
\def\1{\bm{1}}

\DeclareMathAlphabet{\mathsfit}{\encodingdefault}{\sfdefault}{m}{sl}
\SetMathAlphabet{\mathsfit}{bold}{\encodingdefault}{\sfdefault}{bx}{n}

\usepackage{hyperref}
\usepackage{url}
\usepackage{graphicx}
\usepackage{booktabs}
\usepackage{multirow}
\usepackage[table]{xcolor}
\usepackage{adjustbox}
\usepackage{makecell}
\usepackage{caption}
\usepackage{tabularx}
\usepackage{listings}
\usepackage{subcaption}

\lstdefinestyle{prompt}{
    basicstyle=\ttfamily\scriptsize,
    breaklines=true,
    breakatwhitespace=false,
    columns=fullflexible,
    keepspaces=true,
    frame=single,
    showstringspaces=false,
    aboveskip=4pt,
    belowskip=4pt
}

\def\eg{\emph{e.g.}}
\def\ie{\emph{i.e.}}

\title{DMC-5: Dual-Model Collaboration with Cache-aware Context Compression for Long-horizon Agents}

\author{
Haochen Li\thanks{Equal contribution.}, \quad
Xiaobing Tu\footnotemark[1], \quad
Yinggui Wang\thanks{Corresponding author.}, \quad
Jinkui Ren, \quad
Xiantao Zhang
\\
Alibaba Group
}

\iclrfinalcopy % Uncomment for camera-ready version, but NOT for submission.
\begin{document}

\maketitle

\fancyhead{} % disable head line

\vspace{-15pt}
\begin{abstract}

Long-horizon agents repeatedly feed an ever-growing interaction trajectory to a specialized agent model. 
Prefix caching reuses computation over repeated history, but cannot prevent context growth or remove information irrelevant to the current decision. 
This is especially costly when the specialized agent lacks the cached-input discounts available to general-purpose models, making trajectory compression a natural alternative. 
However, existing compression methods on selective context risk losing global information, recursively rewriting summary leads to semantic drift, while repeatedly recompressing the full history shifts the growing cost to the compressor.
Therefore, we introduce \textbf{DMC-5} (Dual-Model Collaboration with CaChe-aware Context Compression), an asymmetric framework that assigns full-history integration to a general compression model and action generation to a specialized agent model.
DMC-5 maintains the raw trajectory as an Append-only Canonical View, preserving global access while allowing the compressor to reuse a stable historical prefix instead of recursively rewriting or recomputing prior context. 
The agent operates only on a compact Decision View with recent raw interactions, reducing repeated input while retaining the information needed for the next action. 
DMC-5 further incorporates \textbf{YAMATO} (integritY-Aware Memory Abstractions for Tool Operations) to preserve execution-critical information during lossy compression.
Experiments on AppWorld, OfficeBench, and 8-objective QA show that DMC-5 reduces agent model input by 38.5$\sim$57.0$\%$ and end-to-end cost by 14.5$\sim$28.0$\%$ with 78.0$\sim$85.0$\%$ compressor cache-hit rates, while improving task performance by 1.0$\sim$5.8$\%$ over the uncompressed baseline.

\end{abstract}

\vspace{-10pt}
\section{Introduction}
\label{sec:introduction}
\vspace{-5pt}

\begin{figure}[t]
\centering
\includegraphics[width=1.0\columnwidth]{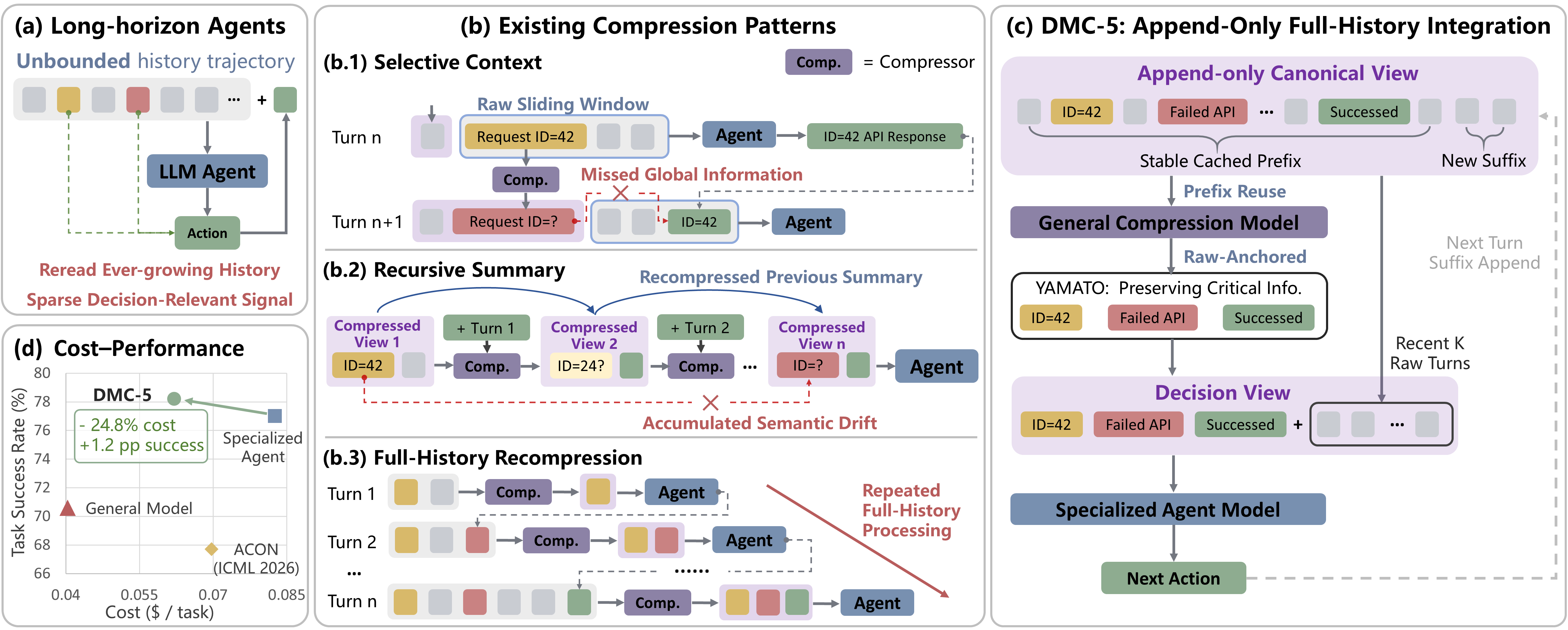}
\caption{Motivation and overview of DMC-5.
(a) Long-horizon agents repeatedly reread a growing trajectory, increasing cost while decision-relevant information is sparse.
(b) Existing compression patterns trade off global information, semantic stability, and repeated compressor processing.
(c) DMC-5 separates full-history integration from action generation: an Append-only Canonical View enables stable prefix reuse for the general compressor, while a compact Decision View retains recent and execution-critical information for the specialized agent.
(d) DMC-5 improves the cost--performance trade-off over the uncompressed specialized agent and existing compression baselines.}
\label{fig-intro}
\vspace{-25pt}
\end{figure}

Long-horizon, tool-using agents repeatedly ask a specialized decision model to reread an ever-growing trajectory of reasoning, actions, and observations~\citep{yao2023react, packer2023memgpt}. 
As shown in Fig.~\ref{fig-intro}(a), an LLM agent repeatedly rereads a growing history, although only a small fraction is relevant to current decision, increasing cost and distracting the model~\citep{laban2025lost}.
Prefix caching reuses stable prefixes~\citep{openai2024promptcache, song2026capc} to reduce computation, but it neither shortens the context length nor removes irrelevant content. 
This cost is especially pronounced for fine-tuned or customized agent endpoints, for which cached-input discounts are often unavailable or limited than those offered for general models.~\citep{openai2024promptcache, google2026tunedcache, azure2026finetunecache, alibaba2026contextcache}.
To address this issue, trajectory compression shifts history integration to a lower-cost compressor, reducing the context repeatedly processed by the agent model.

Existing compressors reduce agent-side context through three update patterns in Fig.~\ref{fig-intro}(b), but each incurs extra end-to-end cost.
\textbf{Selective Context}~\citep{xiao2026agentdiet,ning2026agentomit,hu2024hiagent,ye2026agentfold,verma2026focus} keeps raw trajectory in sliding window and compresses context outside, reducing compression cost at the risk of missing global information and triggering repeated steps.
\textbf{Recursive Summary}~\citep{zhou2025mem1,yu2025memagent,wu2025resum} updates a previous summary with new interactions to avoid full-history rereading, but rewritten summary can accumulate semantic drift~\citep{kang2025acon, shi2025rememr1} and limit prefix-cache reuse.
\textbf{Full-history Recompression}~\citep{li2023selectivecontext,jiang2023longllmlingua,xu2023recomp} rebuilds context from retained raw history, preserving global access but repeatedly processing the growing trajectory.
Overall, reducing the agent-side context alone does not guarantee lower end-to-end cost, as compressor cache misses and extra agent steps caused by information loss can offset the savings.
Taken together, these trade-offs show that token compression ratio alone does not determine end-to-end cost, as compressor cache misses and extra agent steps from information loss may offset agent-side token savings.

Therefore, a practical compression framework must control both compression overhead and the interaction overhead introduced by information loss. 
This requires the compressor and the agent to consume different views of the trajectory.
An ideal compressor should retain access to the full trajectory so that previously observed information remains recoverable, while extracting the context most relevant to the current task.
This full-history role can be assigned to a general model with strong cross-task generalization and discounted cached-input pricing, using an append-only trajectory view to preserve prior information and maintain stable prefix reuse across compression rounds.
The agent, in contrast, should specialize in the downstream task and action space, often through fine-tuning, while consuming only a compact Decision View that keeps information critical for the next action, especially when its deployment provides no or limited cached-input discount.
Together, this asymmetric design assigns full-history integration to a general compression model and compact action generation to specialized agent model, reducing both compression and interaction overhead.

Motivated by this, we introduce \textbf{DMC-5} (Dual-Model Collaboration with CaChe-Aware Context Compression), an asymmetric framework that assigns full-history integration to a general compression model and action generation to a specialized agent model.
Its compressor reads an \textbf{Append-only Canonical View} containing the full-history raw trajectory, so each compression integrates the full history without recursively rewriting previous summaries, while the unchanged prefix remains reusable across compression rounds.
The agent receives a \textbf{Decision View} composed of compact context and recent raw interactions, reducing repeated input while preserving critical information for the next action.
To protect execution-critical state during lossy compression, DMC-5 incorporates \textbf{YAMATO} (integritY-Aware Memory Abstractions for Tool Operations).
YAMATO lets the compressor reference selected raw actions, tool results, and work-product updates through pinned pointers, and deterministically restores the information from the raw records rather than being rewritten by the model.
It also maintains a persistent failure ledger and provenance tags across compression.
Therefore, DMC-5 preserves full-history access during compression, keeps the agent context compact, and protects state whose corruption or omission could affect subsequent execution.

Across AppWorld, OfficeBench, and 8-objective QA, DMC-5 reduces agent-model input by 38.5$\sim$57.0$\%$ and end-to-end cost by 14.5$\sim$28.0$\%$, while achieving 78.0$\sim$85.0$\%$ compressor cache-hit rates. Compared with the uncompressed baseline, task performance improves by 1.0$\sim$5.8$\%$ across the three benchmarks.
In summary, we conclude our contributions as follows:

\begin{itemize}
    \item  We formulate trajectory compression as an end-to-end cost problem, jointly accounting for agent input, compressor overhead, prefix-cache reuse, and interaction length.
    \item We propose DMC-5, separating cache-aware context compression from specialized action generation through Append-only Canonical and compact Decision Views, with YAMATO preserving execution-critical information via deterministic raw-anchored restoration. 
    \item  We evaluate DMC-5 across three long-horizon agent benchmarks.
    It reduces agent input by 38.5$\sim$57.0$\%$ and end-to-end cost by 14.5$\sim$28.0$\%$, reaches 78$\sim$85$\%$ compressor cache hit, and improves task performance by 1.0$\sim$5.8$\%$ over No Compression baseline.
\end{itemize}

\vspace{-5pt}

\section{Method}
\label{sec:method}
\vspace{-5pt}

We propose \textbf{DMC-5} (Dual-Model Collaboration with CaChe-Aware Context Compression) to reduce the end-to-end cost of long-horizon agents by separating full-history integration from action generation, and \textbf{YAMATO} (integritY-Aware Memory Abstractions for Tool Operations) to preserve execution-critical information during lossy compression.
We first formalize the end-to-end cost, then present the dual-model workflow, the two trajectory views, and YAMATO.

% To reduce the end-to-end cost of long-horizon agents, we propose DMC-5 (Dual-Model Collaboration with CaChe-Aware Context
% Compression), an asymmetric framework that separate full-history integration from action generation using two fixed model roles and two trajectory views.
% As shown in Fig.~\ref{fig:method}(a), a general compression model reads the complete \textbf{Append-only Canonical View} and periodically produces a compact state, while a specialized agent acts on a \textbf{Decision View} composed of this  state and recent raw interactions.
% Meanwhile, \textbf{YAMATO} is introduced to preserve execution-critical state during the lossy compression.
% Overall, DMC-5 preserves full-history access and stable prefix reuse for the compressor while reducing the context repeatedly processed by the agent.
% We first formalize the end-to-end cost, then describe the dual-model workflow, the two trajectory views, and YAMATO.

\vspace{-5pt}
\subsection{Problem Setup and End-to-End Cost}
\label{sec:method_setup}

Given a task instruction $q$, let the interaction trajectory after $t$ steps be
\begin{equation}
    \tau_t=(\xi_1,\ldots,\xi_t), \qquad \xi_i=(a_i,o_i),
    \label{eq:trajectory}
\end{equation}
where $a_i$ is the agent action and $o_i$ is the resulting observation.
A full-history agent $F_A$ receives
\begin{equation}
    X_t^{\mathrm{full}}=s_A\oplus q\oplus\tau_t,
    \qquad
    a_{t+1}= F_A(X_t^{\mathrm{full}}),
    \label{eq:full_agent}
\end{equation}
where $s_A$ is the agent system prompt and $\oplus$ denotes serialization and concatenation.
Because each decision conditions on all previous interactions, an interaction recorded at step $i$ is repeatedly included in the inputs of steps $i,i+1,\ldots,T-1$.
Therefore, the cumulative agent input is  $\sum_{t=0}^{T-1} I(X_t^{\mathrm{full}})$, which repeatedly charges for overlapping history as the trajectory grows.

The monetary cost of these overlapping tokens depends on prefix-cache pricing.
For an endpoint $m$ with input $x$ and output $y$, let $I(x)$ and $O(y)$ denote the input and output token counts.
We split the input into provider-reported cached tokens $H_m(x)$ and uncached tokens
$I(x)-H_m(x)$.
The billed cost of one call is
\begin{equation}
    \mathcal{B}_m^{\mathrm{disc}}(x,y)=
    (I(x)-H_m(x))p_m^{\mathrm{miss}}
    +H_m(x)p_m^{\mathrm{hit}}
    +O(y)p_m^{\mathrm{out}},
    \label{eq:call_bill}
\end{equation}
where $p_m^{\mathrm{miss}}$, $p_m^{\mathrm{hit}}$, and $p_m^{\mathrm{out}}$
are the uncached-input, cached-input, and output-token prices.

In practical settings, the agent is specialized for the downstream task through fine-tuning, while many current model providers do not expose discounted cached-input pricing for such customized endpoints~\citep{alibaba2026contextcache}.
Therefore, we target deployments in which the specialized agent receives no cached-input discount, \ie,
$p_m^{\mathrm{hit}}=p_m^{\mathrm{miss}}=p_m^{\mathrm{in}}$:
\begin{equation}
    \mathcal{B}_m^{\mathrm{flat}}(x,y)=
I(x)p_m^{\mathrm{in}}+O(y)p_m^{\mathrm{out}}.
    \label{eq:no_cache_discount}
\end{equation}
Thus, repeated history remains fully billed for the specialized agent, which motivates trajectory compression.
For a generic trajectory compression method, let $X_t^{U}$ denote the compressed input provided to the agent $U$ at step $t$, and let $X_r^{V}$ denote the input of compressor $V$ at compression round $r\in\mathcal{R}$.
Compared with the full-history agent, compression reduces the agent input but introduces an additional compressor cost.
The end-to-end costs are
\begin{equation}
\mathcal{C}_{\mathrm{full}}
=
\sum_{t=0}^{T_{\mathrm{full}}-1}
\mathcal{B}_U^{\mathrm{flat}}(X_t^{\mathrm{full}},y_t^{\mathrm{full}}),
\qquad
\mathcal{C}_{\mathrm{cmp}}
=
\sum_{t=0}^{T_{\mathrm{cmp}}-1}
\mathcal{B}_U^{\mathrm{flat}}(X_t^{U},y_t^{U})
+
\sum_{r\in\mathcal{R}}
\mathcal{B}_V^{\mathrm{disc}}(X_r^{V},M_r),
\label{eq:run_cost}
\end{equation}
where $V$ and $U$ denote the compressor and agent endpoints, $M_r$ is the compressor output, and $T_{\mathrm{full}}$ and $T_{\mathrm{cmp}}$ are the total steps of agent run.

Compression is cost-effective only when the reduction in agent-side cost exceeds the added compressor cost and any interaction overhead caused by information loss.
The compressor cost further depends on prefix-cache reuse: a larger $H_V(X_r^{V})$ moves more input from the uncached price $p_V^{\mathrm{miss}}$ to the lower cached price $p_V^{\mathrm{hit}}$.
Meanwhile, compression may also change agent actions, thereby $T_{\mathrm{cmp}}$ may not equal $T_{\mathrm{full}}$.
We therefore measure end-to-end saving over the complete agent runs as
\begin{equation}
    \Delta\mathcal{C}
    =
    \mathcal{C}_{\mathrm{full}}
    -
    \mathcal{C}_{\mathrm{cmp}},
    \label{eq:cost_saving}
\end{equation}
which accounts for agent-side token reduction, compressor processing, prefix-cache reuse, and changes in interaction length, rather than estimating  compression ratio alone.

\begin{figure*}[t]
    \centering
    \includegraphics[width=\textwidth]{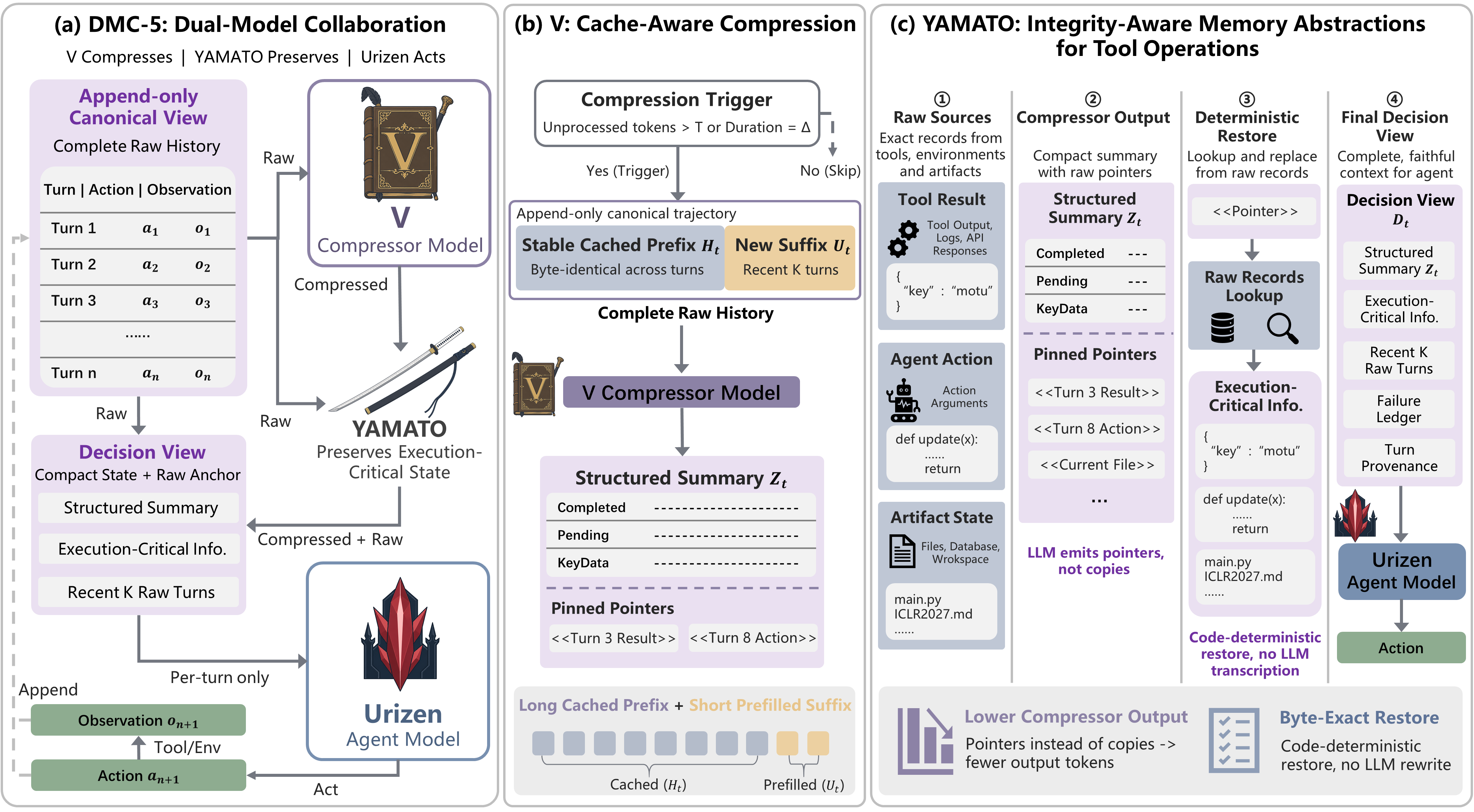}
    \vspace{-5pt}
    \caption{\textbf{Overview of DMC-5.}
    (a) A general compressor V integrates Append-only Canonical View, while a specialized agent Urizen acts on compact Decision View.
    (b) Each compression round reuses the stable historical prefix and only prefills new suffix.
    (c) YAMATO restores execution-critical records through pointers from raw records and combines them with recent raw turns.}
    \label{fig:method}
    \vspace{-15pt}
\end{figure*}

\subsection{DMC-5: Dual-Model Collaboration}
\label{sec:dual_view}

Full-history integration and action generation impose different context requirements: the former benefits from complete historical access, while the latter should avoid repeatedly processing irrelevant history.
Therefore, DMC-5 assigns full-history integration to a \textbf{general compressor} (denoted as \textbf{V}) and action generation to a \textbf{specialized agent} (denoted as \textbf{Urizen}), as shown in Fig.~\ref{fig:method}(a).
Accordingly, the two models consume different views of the same trajectory: an \textbf{Append-only Canonical View} for V and a compact \textbf{Decision View} for Urizen.

The Append-only Canonical view stores the complete raw trajectory and is never replaced by compressed summary.
First, at agent turn $t$, when compression round is triggered, DMC-5 selects the trajectory prefix $\tau^{pre}_{t,K}=(\xi_1,\ldots,\xi_{t-K})$ for compression, while retaining the most recent $K$ turns $\tau^{suf}_{t,K}=(\xi_{t-K+1},\ldots,\xi_t)$ verbatim in the Decision View.
At each compression round, the entire eligible prefix is read directly from the canonical view and sent to the V compressor $F_V$:
\begin{equation}
    X_{t}^{V}=s_V\oplus q\oplus \tau^{pre}_{t,K}, \qquad
    \widehat{M}_t=F_V(X_t^{V}),
    \label{eq:canonical_view}
\end{equation}
where $s_V$ is the compressor prompt.
Thus, V constructs each compressed state $\widehat{M}_t$ directly from the raw trajectory, preserving access to earlier information while avoiding recursive summary drift.

Then, DMC-5 applies YAMATO operation $\mathcal{Y}$ to the output of V compressor to restore execution-critical state from the Append-only Canonical View:
\begin{equation}
    M_t=\mathcal{Y}(\widehat{M}_t;\tau^{pre}_{t,K})\oplus\tau^{suf}_{t,K}.
\end{equation}

After that, the Decision View consists of agent prompt $s_U$, task instruction $q$ and context $M_t$: 
\begin{equation}
    X_t^{U}
    =
    s_U\oplus q\oplus M_{t}, \qquad a_{t+1}= F_U(X_t^{U})
    \label{eq:decision_view}
\end{equation}
where $F_U$ is the Urizen agent.
Thus, Urizen reads older history through $M_r$ while retaining recent interactions verbatim.
Subsequent actions $a_{t+1}$ and observations $o_{t+1}$ are appended only to canonical view as the suffix.

With this dual-model design, DMC-5 preserves full-history access for V while providing Urizen with only the compact state and recent raw interactions needed for action generation, reducing repeated agent-side context.

\subsection{V: Cache-Aware Context Compression}
\label{sec:cache_integration}
\vspace{-5pt}

Full-history compression preserves access to earlier information, but repeatedly processing the growing trajectory can shift the cost from the agent to the compressor.
V addresses this cost through append-only prefix reuse, as shown in Fig.~\ref{fig:method}(b). 
Compression is triggered when the accumulated uncompressed history exceeds a token threshold $T$ or the compression interval $\Delta$ is reached.
At agent turn $t$, V's input is decomposed as
\vspace{-5pt}
\begin{equation}
    X_t^{V}=H_t\oplus U_t,
    \label{eq:append_only}
    \vspace{-5pt}
\end{equation}
where $H_t$ is the previously processed prefix and $U_t$ contains newly eligible turns since the last compression.
Because the canonical trajectory is append-only, $H_t$ remains byte-identical and in the same order across compression events, allowing the endpoint to serve $H_t$ at the cached-input price while only $U_t$ is prefilled as new input.
Therefore, V retains access to the full history without repeatedly paying the uncached-input cost for the same trajectory.

Once the compression is triggered, V outputs a structured summary containing completed and pending progress, key data, and pinned pointers to execution-critical raw records in canonical view.
The output is passed to YAMATO and is never written back to the canonical trajectory.
With the cache-aware context compression design, V retains full-history access without recursive summary, while keeping the historical prefix reusable across compression rounds.

\vspace{-10pt}
\subsection{YAMATO: Integrity-Aware Memory Abstractions for Tool Operations}
\label{sec:yamato}
\vspace{-5pt}

Lossy compression reduces agent input, but corrupted or omitted tool operation state can cause incorrect or repeated actions. 
Tool operation often depends on exact identifiers, arguments, returned values, and artifact states.
As shown in Fig.~\ref{fig:method}(c), YAMATO preserves such execution-critical information through deterministic restoration rather than model transcription, maintaining information integrity while reducing output tokens.

Specifically, YAMATO treats records whose values may affect subsequent action (\eg, identifiers, arguments, returned values, and work-product
states) as execution-critical, and lets compressor reference them through pointers.
In particular, V outputs a structured summary $Z_t$ and a set of pinned pointers $\Pi_t$:
\begin{equation}
    \widehat{M}_t=(Z_t,\Pi_t).
\end{equation}
YAMATO $\mathcal{Y}$ then restores the referenced canonical records deterministically,
maintains persistent failure state, and attaches turn provenance to the
semantic summary:
\begin{equation}
    M_t = \mathcal{Y}(\widehat{M}_t;\tau^{pre}_{t,K})\oplus\tau^{suf}_{t,K}
    =
    \overline{Z}_t
    \oplus
    \mathcal{R}(\Pi_t;\tau^{pre}_{t,K})
    \oplus
    L_t
    \oplus
    \tau^{suf}_{t,K},
    \label{eq:yamato_restore}
\end{equation}
where $\overline{Z}_t$ denotes the provenance-aware semantic summary, $\mathcal{R}$ denotes deterministic restoration operation, $L_t$ retains failed actions across compression rounds.
Detailed preservation rules, pointer schemas, and provenance handling are
provided in Appendix~\ref{app:yamato_details}.

In conclusion, YAMATO keeps the semantic state compact while restoring selected execution-critical information byte-for-byte from the canonical trajectory.
It limits information loss that could otherwise increase interaction steps, while pointers reduce compressor output tokens and cost.

Overall, each component of DMC-5 corresponds directly to the end-to-end cost in Eq.~\ref{eq:run_cost}.
The Decision View reduces the agent input $X^{U}$, append-only compression increases cached reuse within the compressor input $X^{V}$, and YAMATO limits additional interaction steps $T_{\mathrm{cmp}}$ caused by information loss while reducing compressor output $M$ through pinned pointers.
DMC-5 therefore targets both terms of $\mathcal{C}_{\mathrm{cmp}}$ rather than optimizing the agent-side compression ratio alone.

\vspace{-5pt}
\section{Experiments}
\vspace{-5pt}

\label{sec:experiments}

We evaluate four aspects of DMC-5: 1) end-to-end quality--cost trade-off, 2) cache-aware context compression, 3) asymmetric dual-model collaboration, 4) execution-critical information integrity.
\vspace{-8pt}
\subsection{Experimental Setup}
\label{sec:experimental_setup}
\vspace{-5pt}

\textbf{Benchmarks.}
We evaluate DMC-5 on three long-horizon agent benchmarks.
\textbf{AppWorld}~\citep{trivedi2024appworld} contains 168 executable multi-step application tasks, while \textbf{OfficeBench}~\citep{wang2024officebench} contains 192 shorter office-oriented tool-use tasks; we report accuracy on both.
\textbf{8-objective QA} contains 200 multi-objective tasks with substantially longer accumulated context, following~\citep{kang2025acon} and report EM/F1.
Together, they cover complex executable interaction, shorter workflows, and long-context information integration.
Detailed task splits are provided in Appendix~\ref{app:benchmark_protocols}.
All results are averaged over three independent runs, reported in Appendix~\ref{sec:stability}

\textbf{Models and baselines.}
For fair comparison, we use the same fine-tuned Qwen3.5-27B as the agent and DeepSeek-V4-flash as the compressor, comparing against \textsc{No Compression}, \textsc{FIFO}~\citep{yang2024swe}, \textsc{AgentDiet}~\citep{xiao2026agentdiet}, and \textsc{ACON}~\citep{kang2025acon}.
ACON is evaluated under official implementation, while AgentDiet is reproduced since no official code is available, reported in Appendix~\ref{app:baseline_implementation}.
Other model configurations are specified in the corresponding ablations.

\textbf{Cost billing.}
We compute billed cost from observed uncached input, cached input, and output tokens using Eq.(\ref{eq:run_cost}).
All reported costs come from total tasks run and include both agent and compressor calls.
Cache Hit denotes the provider-reported cached tokens.
Details please refer to Appendix~\ref{app:model_pricing}

\textbf{Hyperparameters.}
We use $K=1$, $\Delta=3$, and $T=6000$ by default.
The compressor uses temperature $0$ with thinking disabled, while the agent follows benchmark-default decoding.
Please see Appendix~\ref{app:inference_config} for more details.

\vspace{-5pt}
\subsection{Comparison with Existing Compression Methods}
\label{sec:overall_results}
\vspace{-5pt}

\begin{table*}[t]
  \centering
  \caption{Overall comparison across three agent benchmarks. 
  Score changes are absolute differences from No Compression; input and cost changes are relative.
  Agent Input is averaged per task, and Cache Hit reports compressor prefix-cache reuse rate.
  }
  \label{tab:main_results}
  \vspace{-10pt}
  \small
  \setlength{\tabcolsep}{3.6pt}
  \renewcommand{\arraystretch}{1.05}
  \begin{adjustbox}{max width=\textwidth}
  \begin{tabular}{llllrrrrr}
    \toprule
    \multirow{2}{*}{Benchmark} &
    \multirow{2}{*}{Method} &
    \multirow{2}{*}{Metric} & \multirow{2}{*}{Score$\uparrow$} &
    \multicolumn{2}{c}{Agent Input $\downarrow$} & Compressor &
    \multicolumn{2}{c}{Total Cost $\downarrow$} \\
    
    \cmidrule(lr){5-6}
    \cmidrule(lr){8-9}
    & &  &  & Tokens & Change & Cache Hit $\uparrow$ & Total & Change \\
    \midrule

    \multicolumn{9}{c}{
      \textbf{Agent Model:} Fine-tuned Qwen3.5-27B
      \quad \quad \quad \quad \quad \quad
      \textbf{Compressor Model:} DeepSeek-V4-flash
    } \\
    \midrule

    % ==================== AppWorld ====================
    \multirow{4}{*}{AppWorld}
      & \cellcolor{gray!15}No Compression
      & \cellcolor{gray!15}Acc.
      & \cellcolor{gray!15}77.0
      & \cellcolor{gray!15}258,996
      & \cellcolor{gray!15}--
      & \cellcolor{gray!15}--
      & \cellcolor{gray!15}13.71
      & \cellcolor{gray!15}-- \\

      & FIFO~\citep{yang2024swe}
      & Acc.
      & 36.3 {\scriptsize($-40.7$)}
      & 155,335
      & -40.02$\%$
      & --
      & \textbf{8.98}
      & \textbf{-34.50$\%$} \\

      & AgentDiet~\citep{xiao2026agentdiet}
      & Acc.
      & 64.8 {\scriptsize($-12.2$)}
      & 181,082
      & -30.08$\%$
      & 0.0
      & 11.93
      & -12.98$\%$ \\

      & ACON~\citep{kang2025acon}
      & Acc.
      & 67.3 {\scriptsize($-9.7$)}
      & 160,684
      & -37.96$\%$
      & 0.0
      & 11.57
      & -15.61$\%$ \\

      & \cellcolor{blue!8}\textbf{DMC-5}
      & \cellcolor{blue!8}Acc.
      & \cellcolor{blue!8}\textbf{78.2}
        {\scriptsize\textcolor{red}{($+1.2$)}}
      & \cellcolor{blue!8}\textbf{138,431}
      & \cellcolor{blue!8}\textbf{-46.55$\%$}
      & \cellcolor{blue!8}\textbf{80.3}$\%$
      & \cellcolor{blue!8}10.31
      & \cellcolor{blue!8}-24.80$\%$ \\

    \midrule

    % ==================== OfficeBench ====================
    \multirow{4}{*}{OfficeBench}
      & \cellcolor{gray!15}No Compression
      & \cellcolor{gray!15}Acc.
      & \cellcolor{gray!15}87.0
      & \cellcolor{gray!15}76,868
      & \cellcolor{gray!15}--
      & \cellcolor{gray!15}--
      & \cellcolor{gray!15}5.25
      & \cellcolor{gray!15}-- \\

      & FIFO~\citep{yang2024swe}
      & Acc.
      & 59.3 {\scriptsize($-27.7$)}
      & 53,281
      & -30.6$\%$
      & --
      & \textbf{3.95}
      & \textbf{-24.80$\%$} \\

      & AgentDiet~\citep{xiao2026agentdiet}
      & Acc.
      & 83.8 {\scriptsize($-3.2$)}
      & 55,328
      & -28.02$\%$
      & 0.0
      & 4.13
      & -21.33$\%$ \\
      
      & ACON~\citep{kang2025acon}
      & Acc.
      & 83.3 {\scriptsize($-3.7$)}
      & 49,447
      & -35.67$\%$
      & 0.0
      & 4.19
      & -20.19$\%$ \\

      & \cellcolor{blue!8}\textbf{DMC-5}
      & \cellcolor{blue!8}Acc.
      & \cellcolor{blue!8}\textbf{88.0}
        {\scriptsize\textcolor{red}{($+1.0$)}}
      & \cellcolor{blue!8}\textbf{47,307}
      & \cellcolor{blue!8}\textbf{-38.46$\%$}
      & \cellcolor{blue!8}\textbf{78.0$\%$}
      & \cellcolor{blue!8}4.49
      & \cellcolor{blue!8}-14.48$\%$ \\

    \midrule

    % ==================== 8-objective QA ====================
    \multirow{5}{*}{8-objective QA}

      & \cellcolor{gray!15}No Compression
      & \cellcolor{gray!15}\shortstack[l]{EM\\[1pt]F1}
      & \cellcolor{gray!15}\shortstack[l]{39.1\\[1pt]47.6}
      & \cellcolor{gray!15}560,556
      & \cellcolor{gray!15}--
      & \cellcolor{gray!15}--
      & \cellcolor{gray!15}35.40
      & \cellcolor{gray!15}-- \\

      % &
      % & \multicolumn{7}{c}{\multirow{2}{*}{\textit{Fail}}} \\
      % & FIFO~\citep{yang2024swe}
      % & \multicolumn{7}{c}{} \\

      & AgentDiet~\citep{xiao2026agentdiet}
      & \shortstack[l]{EM\\[1pt]F1}
      & \shortstack[l]{
          35.5 {\scriptsize($-3.6$)}\\[1pt]
          43.6 {\scriptsize($-4.0$)}
        }
      & 225,041
      & -59.85$\%$
      & 0.0
      & 25.88
      & -26.89$\%$ \\
      
      & ACON~\citep{kang2025acon}
      & \shortstack[l]{EM\\[1pt]F1}
      & \shortstack[l]{
          34.2 {\scriptsize($-4.9$)}\\[1pt]
          41.5 {\scriptsize($-6.1$)}
        }
      & \textbf{224,119}
      & \textbf{-60.02$\%$}
      & 0.0
      & 25.94
      & -26.72$\%$ \\

      & \cellcolor{blue!8}\textbf{DMC-5}
      & \cellcolor{blue!8}\shortstack[l]{EM\\[1pt]F1}
      & \cellcolor{blue!8}\shortstack[l]{
          \textbf{44.9} {\scriptsize\textcolor{red}{($+5.8$)}}\\[1pt]
          \textbf{54.0} {\scriptsize\textcolor{red}{($+6.4$)}}
        }
      & \cellcolor{blue!8}240,937
      & \cellcolor{blue!8}-57.02$\%$
      & \cellcolor{blue!8}\textbf{85.0$\%$}
      & \cellcolor{blue!8}\textbf{25.50}
      & \cellcolor{blue!8}\textbf{-27.97$\%$} \\

    \bottomrule
  \end{tabular}
  \end{adjustbox}

  % \vspace{2pt}
  % \parbox{\textwidth}{\footnotesize
  %   \textit{Note.} \textit{Fail} means that no task completed because the agent
  %   entered non-terminating interaction loops.
  % }
\vspace{-20pt}
\end{table*}

\noindent\textbf{Overall comparison.}
Table~\ref{tab:main_results} compares DMC-5 with full-history prompting and three representative compressors across workloads of different lengths.
AppWorld emphasizes multi-step executable interaction, OfficeBench contains shorter tool-use trajectories, and 8-objective QA has the largest accumulated input.
Across all three settings, DMC-5 reduces agent input by 38.46$\sim$57.02$\%$ and end-to-end cost by 14.48$\sim$27.97$\%$, while improving task performance by 1.0$\sim$5.8$\%$ over No Compression.
The cost reduction is smallest on the shorter OfficeBench, where compression has less opportunity to amortize its overhead, and largest on 8-objective QA with longer trajectories.

The baselines expose the trade-offs discussed in Section~\ref{sec:introduction}.
FIFO achieves low cost by retaining only recent context, but loses 27.7$\sim$40.7$\%$ accuracy on AppWorld and OfficeBench, and fails on 8-objective QA due to execution-critical information loss and non-terminating interaction loops.
AgentDiet selectively removes locally redundant context, reducing agent input by 28.02$\sim$59.85\% and end-to-end cost by 12.98$\sim$26.89\%, while slightly outperforming ACON on OfficeBench and 8-objective QA.
ACON achieves stronger agent-input reduction on AppWorld and OfficeBench through global compression, but its recursive summaries provide no reusable prefix cache and still reduce task performance by 3.7$\sim$9.7$\%$.
Benefiting from the Append-only Canonical View, DMC-5 retains full-history access and 78.0$\sim$85.0$\%$ cache-hit rate while restricting the agent to a compact Decision View, reaching 78.2$\%$ on AppWorld, 88.0$\%$ on OfficeBench, and 44.9/54.0$\%$ EM/F1 on 8-objective QA, showing better cost saving and task performance for long-horizon agent.

\begin{figure}[t]
\centering
\vspace{-10pt}
\includegraphics[width=1.0\columnwidth]{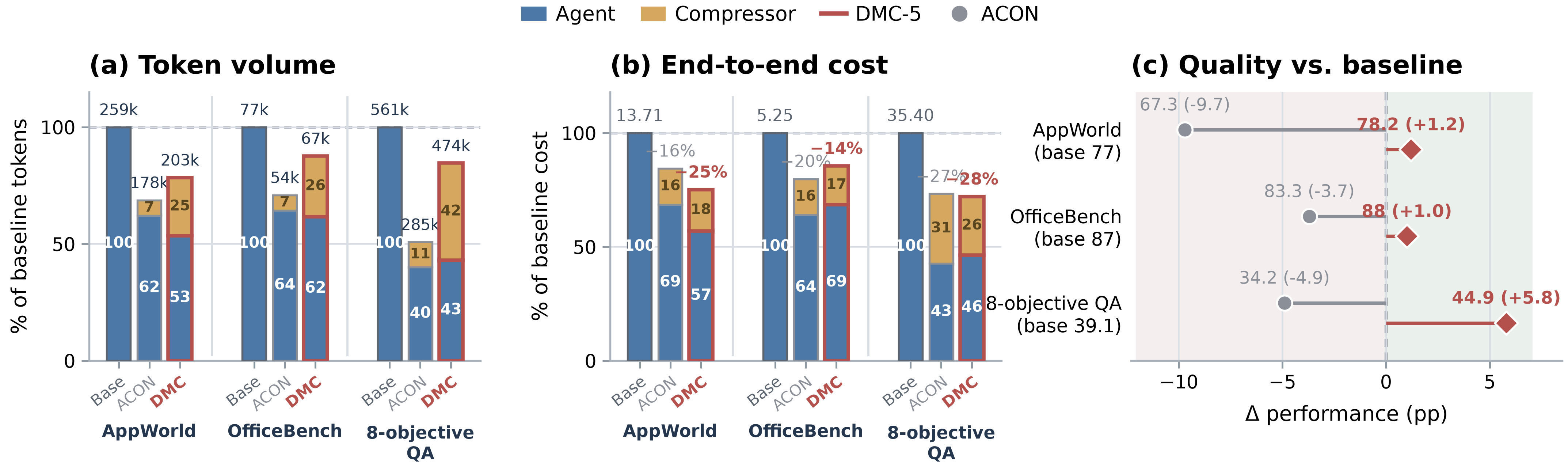}
  \vspace{-20pt}
\caption{Comparison on token, cost and quality across three benchmarks.
(a) Total token volume, decomposed into agent and compressor input.
(b) Billed cost under the same decomposition.
(c) Change in the task performance relative to No Compression.}
\label{fig:tradeoff}
  \vspace{-10pt}
\end{figure}

% \vspace{-10pt}
\noindent\textbf{Token Volume vs. End-to-End Cost.}
Fig.~\ref{fig:tradeoff} compares total token volume, end-to-end cost and task quality.
Although DMC-5 processes more compressor input than ACON because its Canonical View retains full-history access, 78$\sim$85$\%$ of this input is served from cache.
As a result, the larger token volume does not translate into higher compressor cost, and DMC-5 achieves end-to-end cost comparable (+5.7$\%$) to or even lower (-1.25$\sim$-9.19$\%$) than ACON across three benchmarks.
At the same time, DMC-5's Decision View preserves more information for action generation and enhances task performance by 1.0$\sim$5.8$\%$, showing a better quality--cost trade-off by dual-model collaboration with cache-aware context compression.

\vspace{-10pt}
\subsection{Analysis on Cache-Aware Context Compression}
\label{sec:cache_results}
\vspace{-5pt}

\begin{table}[t]
\centering
\caption{Comparison of representative recursive-summary ACON and append-only DMC-5.}
\label{tab:append_only_ablation}
  \vspace{-10pt}
\setlength{\tabcolsep}{3.2pt}
\renewcommand{\arraystretch}{1.15}

\begin{adjustbox}{max width=\textwidth}
\begin{tabular}{llrrrrrr}
\toprule
Benchmark
& Strategy
& \makecell{Cached Comp.\\Tokens}
& \makecell{Uncached Comp.\\Tokens $\downarrow$}
& \makecell{Comp.\\Cost $\downarrow$}
& \makecell{End-to-End\\Cost $\downarrow$}
& \makecell{Cost per\\Success $\downarrow$}
& \makecell{Metric\\Score $\uparrow$} \\
\midrule

\multirow{2}{*}{AppWorld}
& Recursive Summary (ACON)
& 0
& 17,269
& \textbf{2.17}
& 11.57
& 0.1024
& 67.3 \\
& \cellcolor{blue!8}Append-only (DMC-5)
& \cellcolor{blue!8}\textbf{51,837}
& \cellcolor{blue!8}\textbf{12,717}
& \cellcolor{blue!8}2.50
& \cellcolor{blue!8}\textbf{10.31}
& \cellcolor{blue!8}\textbf{0.0781}
& \cellcolor{blue!8}\textbf{78.2} \\
\midrule

\multirow{2}{*}{OfficeBench}
& Recursive Summary (ACON)
& 0
& 5,052
& \textbf{0.83}
& \textbf{4.19}
& \textbf{0.0262}
& 83.3 \\
& \cellcolor{blue!8}Append-only (DMC-5)
& \cellcolor{blue!8}\textbf{15,635}
& \cellcolor{blue!8}\textbf{4,410}
& \cellcolor{blue!8}0.89
& \cellcolor{blue!8}4.49
& \cellcolor{blue!8}0.0266
& \cellcolor{blue!8}\textbf{88.0} \\
\midrule

\multirow{2}{*}{8-objective QA}
& Recursive Summary (ACON)
& 0
& 60,441
& 10.84
& 25.94
& 0.0474
& 34.2 \\
& \cellcolor{blue!8}Append-only (DMC-5)
& \cellcolor{blue!8}\textbf{198,150}
& \cellcolor{blue!8}\textbf{34,844}
& \cellcolor{blue!8}\textbf{9.08}
& \cellcolor{blue!8}\textbf{25.50}
& \cellcolor{blue!8}\textbf{0.0355}
& \cellcolor{blue!8}\textbf{44.9} \\
\bottomrule
\end{tabular}
\end{adjustbox}
\scriptsize
\textit{Note. Recursive Summary} is implemented follows ACON under the same prompting and dual-model structure of DMC-5.

  \vspace{-20pt}
\end{table}

\begin{figure*}[t]
    \centering
    % Left: append-only vs. recursive summary
    \begin{minipage}[t]{0.64\textwidth}
        \centering
    \vspace{-10pt}
        \includegraphics[height=3.3cm]{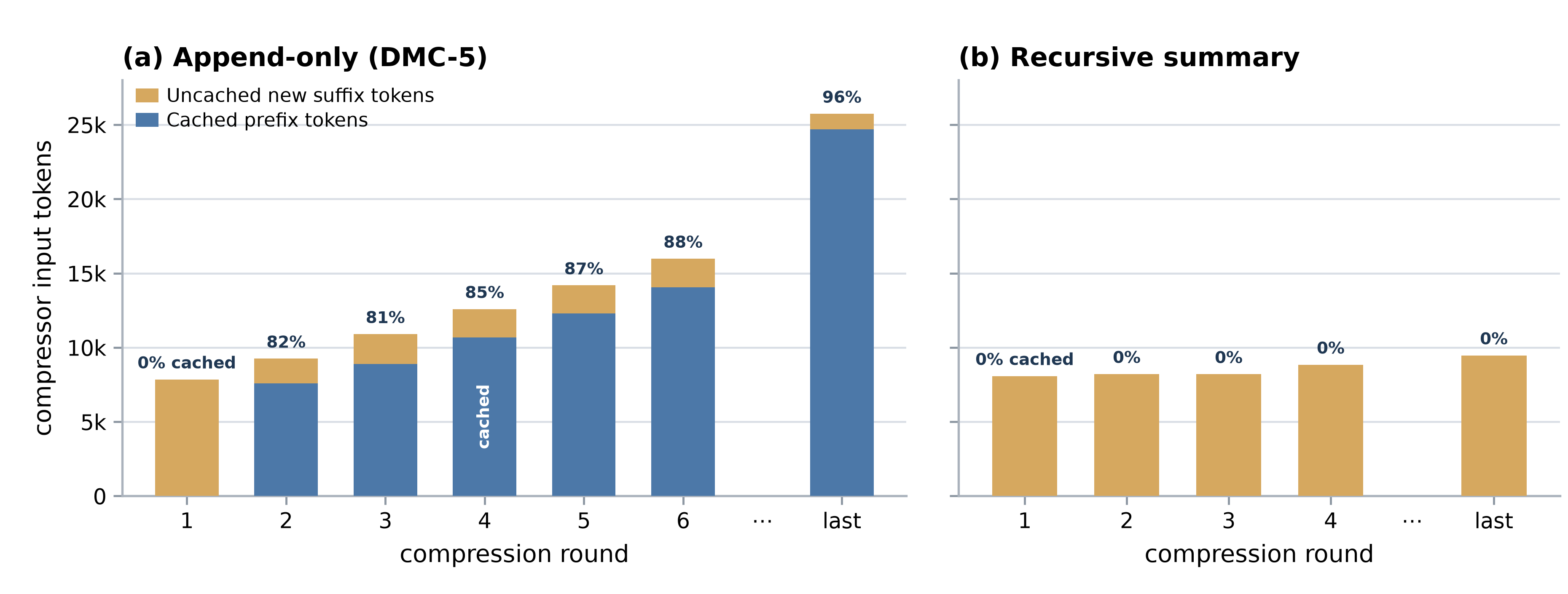}
    \vspace{-10pt}
        \captionof{figure}{
        Compressor cache structure under append-only and recursive summary
        across compression rounds.
        }
    \label{fig:compressor_cache_structure}
    \end{minipage}
    % \hfill
    % Right: compression frequency table
    \begin{minipage}[t]{0.32\textwidth}
        \centering

        \captionsetup{
            type=table,
            font=footnotesize,
            skip=1pt
        }
        \captionof{table}{Effect of compression frequency on OfficeBench.}
        \label{tab:compression_count}

        \footnotesize
        \setlength{\tabcolsep}{2pt}
        \renewcommand{\arraystretch}{0.90}
        \vspace{5pt}
        \begin{tabular}{cccc}
        \toprule
        Comp. 
        & Task 
        & Cost 
        & Acc.  \\
        Count
        & Count
        & Change
        & Change \\
        \midrule
        1       & 75 & +10.7\% & 0 \\
        2       & 27 & -26.5\% & +3.7\% \\
        3--4    & 31 & -20.5\% & +9.7\% \\
        $\geq$5 & 40 & -16.6\% & +7.5\% \\
        \bottomrule
        \end{tabular}
    \end{minipage}
\end{figure*}

\noindent\textbf{Append-only vs. Recursive Summary.}
Table~\ref{tab:append_only_ablation} shows that compared with ACON-style recursive summary, append-only DMC-5 configuration reuses 15.6K$\sim$198.2K cached tokens per task and reduces uncached compressor input by 12.7$\sim$42.3\% across the three benchmarks, while improving task scores by 4.7$\sim$10.9$\%$.
Despite processing more total compressor tokens, DMC-5's compressor cost remains comparable to recursive compression, with 23.7$\%$ and 25.1$\%$ lower cost per success on AppWorld and 8-objective QA, respectively.
Fig.~\ref{fig:compressor_cache_structure} further shows that the cache-hit rate of append-only input increases from 0 to 96$\%$ by the last compression, whereas recursive compression records no cache hits.
Therefore, DMC-5 retains full-history access while leveraging low-cost prefix caching to handle repetitive history, achieving higher task performance and lower end-to-end cost.

%reducing both the end-to-end cost and cost per successful task.

\begin{table*}[t]
\centering
% Left: Net saving + Per-step cost
% =========================================================
\begin{minipage}[t]{0.425\textwidth}
\vspace{-10pt}
\centering

% ---------- Per-step cost figure ----------
\includegraphics[width=0.9\linewidth]{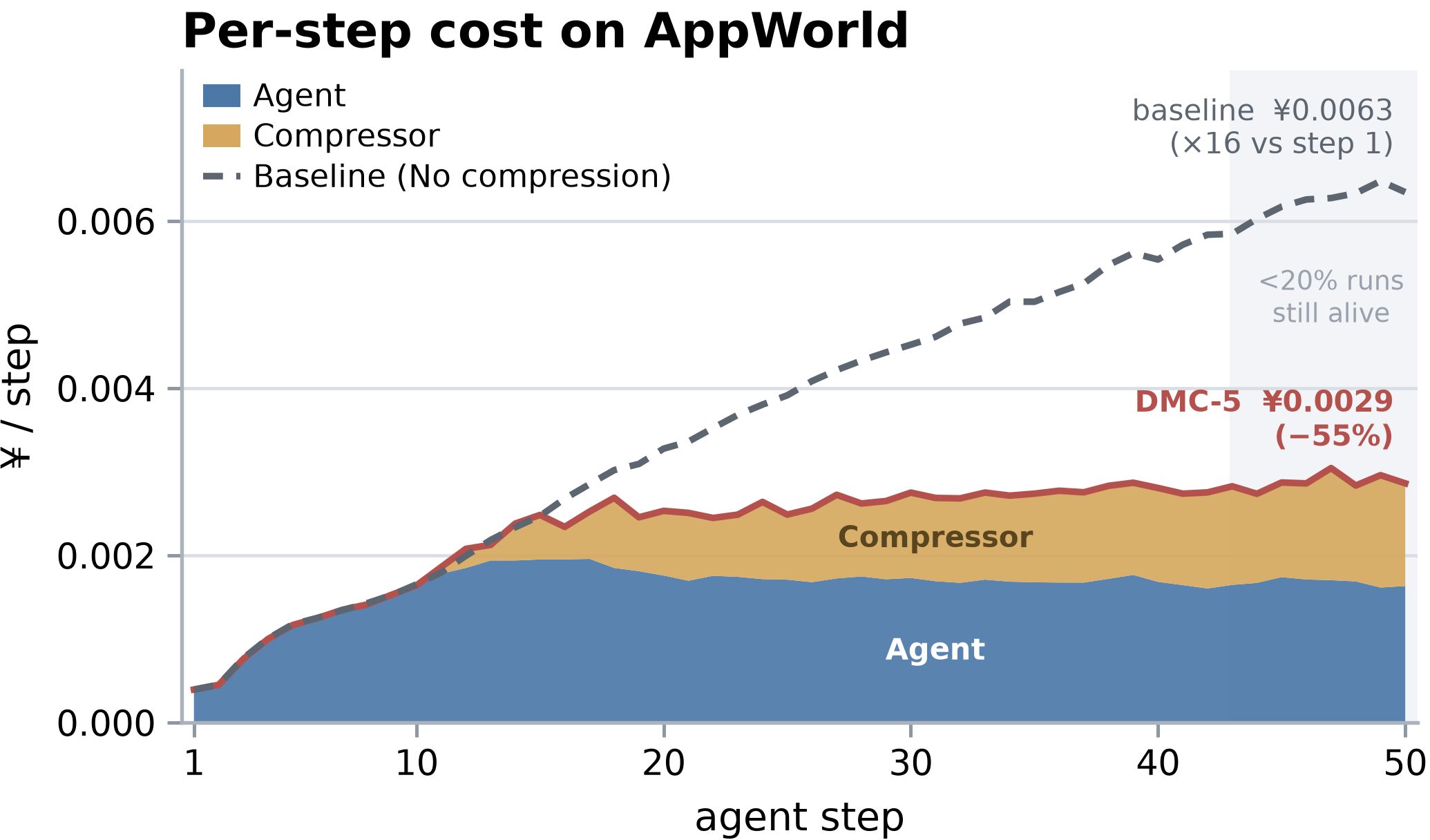}

\vspace{-0.5mm}

\captionsetup{
    type=figure,
    font=footnotesize,
    skip=1pt
}
\captionof{figure}{Per-step cost on AppWorld.}
\label{fig:appworld_cost_per_step}
\vspace{3pt}

% ---------- Net-saving figure ----------
\includegraphics[width=0.87\linewidth]{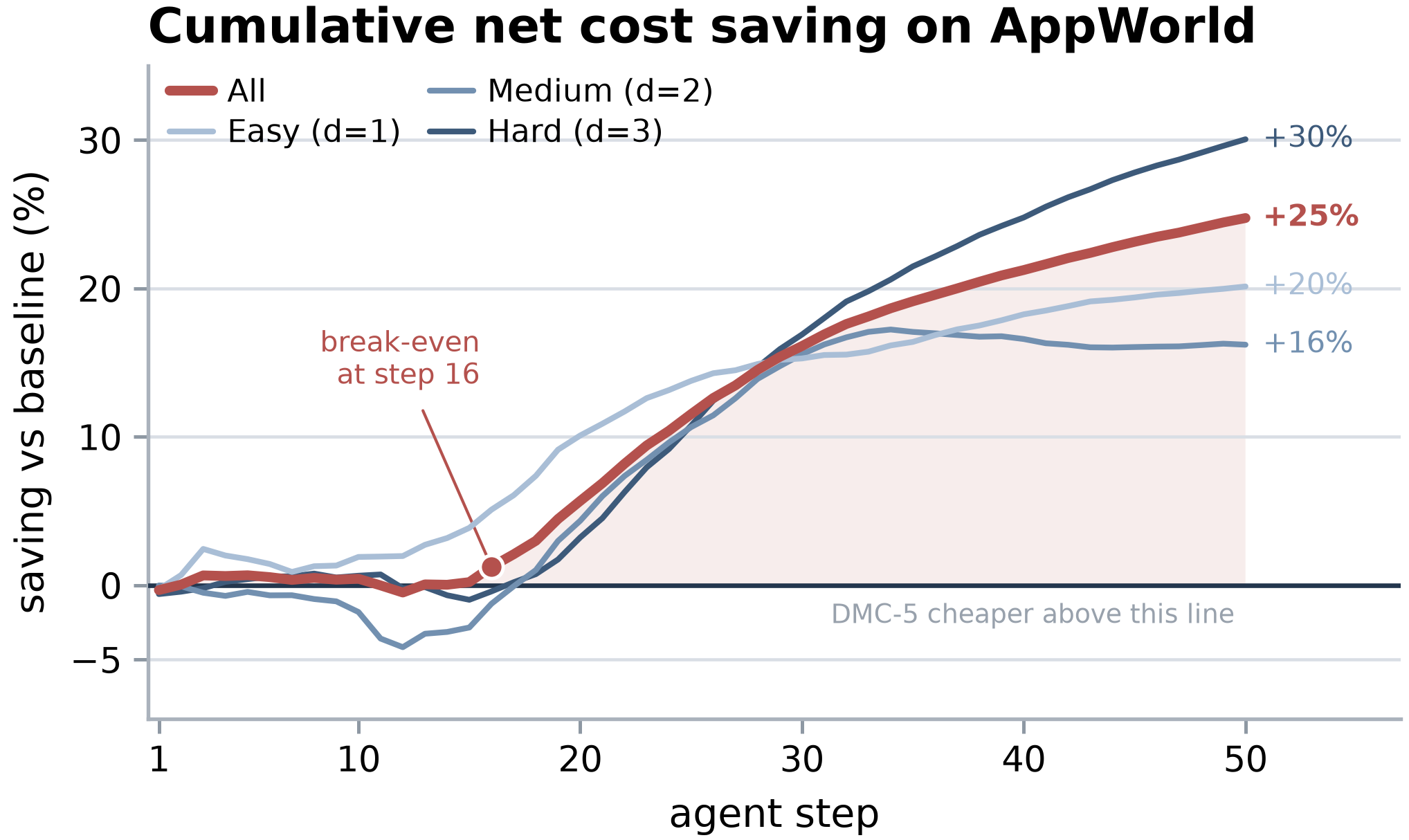}

\vspace{-0.5mm}

\captionsetup{
    type=figure,
    font=footnotesize,
    skip=1pt
}
\captionof{figure}{Net cost saving on AppWorld.}
\label{fig:appworld_net_saving}

\end{minipage}
% =========================================================
% Right: Model-role allocation
% =========================================================
\begin{minipage}[t]{0.55\textwidth}
\vspace{-10pt}
\centering

\caption{Effect of model-role allocation on AppWorld.}
\label{tab:model_role}

\footnotesize
\setlength{\tabcolsep}{2pt}
\renewcommand{\arraystretch}{1.05}

\begin{tabular}{lccc}
\toprule
Setting
& \makecell{Success\\$\uparrow$}
& \makecell{Total \\Cost $\downarrow$}
& \makecell{Cost / \\Success$\downarrow$} \\
\midrule

\multicolumn{4}{l}{\textit{Specialized model: Fine-tuned Qwen3.5-27B}} \\

\cellcolor{gray!15}Specialized only
& \cellcolor{gray!15}77.0\%
& \cellcolor{gray!15}13.71
& \cellcolor{gray!15}0.1051 \\

Specialized $\rightarrow$ Specialized
& 77.4\%
& 21.49
& 0.1634 \\

\midrule
\multicolumn{4}{l}{\textit{General model: DeepSeek-V4-flash}} \\

General only
& 70.6\%
& \textbf{6.93}
& \textbf{0.0591} \\

General $\rightarrow$ General
& 67.9\%
& 9.07
& 0.0796 \\

\cellcolor{blue!8}General $\rightarrow$ Specialized(DMC-5)
& \cellcolor{blue!8}\textbf{78.2\%}
& \cellcolor{blue!8}10.31
& \cellcolor{blue!8}0.0781 \\

\midrule

\multicolumn{4}{l}{\textit{General model: Qwen3.7-flash}} \\

General only
& 67.4\%
& \textbf{4.40}
& \textbf{0.0389} \\

General $\rightarrow$ General
& 61.3\%
& 7.81
& 0.0758 \\

\cellcolor{blue!8}General $\rightarrow$ Specialized(DMC-5)
& \cellcolor{blue!8}\textbf{73.8\%}
& \cellcolor{blue!8}9.50
& \cellcolor{blue!8}0.0766 \\

\bottomrule
\end{tabular}

\vspace{1.5pt}
\footnotesize
\textit{Note.} $A \rightarrow B$ applies $A$ as compressor and $B$ as agent.

\end{minipage}

\vspace{-2mm}
\vspace{-15pt}
\end{table*}

\noindent\textbf{Longer trajectories benefit more from DMC-5.}
Fig.~\ref{fig:appworld_cost_per_step} shows how the cost structure changes as an AppWorld trajectory grows.
Without compression, per-step cost continues to increase with the accumulated history.
After compression, DMC-5 reduces the agent-side cost and eventually drives it downward, indicating that the Decision View removes substantial redundant history, while the Append-only Canonical View keeps the added compressor cost growing slowly through prefix reuse.
Table~\ref{tab:compression_count} shows the same trend that tasks compressed only once cost 10.7$\%$ more than No Compression, whereas tasks with at least two compressions save 16.6$\sim$26.5$\%$ and improve accuracy by 3.7$\sim$9.7$\%$.
As these per-step savings accumulate, Fig.~\ref{fig:appworld_net_saving} shows that DMC-5 reaches break-even around step 16 and achieves 25$\%$ net cost saving by step 50.
Easy tasks realize savings earlier, while hard tasks gain more as the trajectory grows, reaching 30$\%$ cumulative saving.
Overall, longer tasks trigger compression more often and provide more opportunities to reuse both the compact Decision View and the cached prefix of Canonical View, producing larger net savings.

\vspace{-5pt}
\subsection{Analysis on Asymmetric Dual-Model Collaboration}
\label{sec:model_roles}
\vspace{-5pt}

Table~\ref{tab:model_role} evaluates whether history integration and action generation benefit from different model roles.
Using the same compressor, replacing the general agent with the specialized fine-tuned Qwen3.5-27B improves AppWorld success from 67.9$\%$ to 78.2$\%$ with DeepSeek-V4-flash, and from 61.3$\%$ to 73.8$\%$ with Qwen3.7-flash.
This shows that the specialized model remains important for downstream action generation even when it no longer processes the full trajectory.

At the same time, moving full-history integration to a general compressor reduces the cost of specialized calls.
Using the specialized model itself for compression raises total cost to 21.49 with only 77.4$\%$ success, whereas DeepSeek-V4-flash$\rightarrow$Specialized achieves 78.2$\%$ at 10.31.
Qwen3.7-flash further lowers the cost to 9.50, but its lower 73.8$\%$ success shows that the compressor must still preserve an accurate Decision View.
Overall, the results support the asymmetric design: general models provide cheaper full-history integration, while specialized agents preserve action quality.

\vspace{-5pt}
\subsection{Analysis on Execution-Critical Information Integrity}
\label{sec:yamato_results}
\vspace{-5pt}

\noindent\textbf{YAMATO preserves execution-critical information.}
Table~\ref{tab:yamato_main} isolates YAMATO from history organization by applying the same integrity layer to both recursive ACON and append-only DMC-5.
On ACON, YAMATO reduces compressions containing errors from 4.2$\%$ to 1.0$\%$ and corrupted execution-critical items from 1.23$\%$ to 0.35$\%$, improving recall of execution-critical items by 23.8$\%$.
On DMC-5, the corresponding rates drop from 3.0$\%$ to 0.04$\%$ and from 0.33$\%$ to 0.01$\%$, while recall of execution-critical items reaches 98.3$\%$.
The consistent reductions under various compression mechanisms show that deterministic restoration protects execution-sensitive state.

\noindent\textbf{Fewer integrity errors lead to fewer repeated failures.}
YAMATO also reduces repeated failures and improves task completion.
Adding YAMATO to DMC-5 raises task accuracy from 69.1$\%$ to 78.2$\%$, while runs with at least five repeated failed actions fall from 1.6$\%$ to 0.2$\%$ and the maximum same-action repetition drops from 29 to 6.
ACON shows the same trend after adding YAMATO, with success increasing from 67.3$\%$ to 72.0$\%$ and the maximum repetition chain decreasing from 39 to 12.
For DMC-5, the resulting reduction in repeated interaction also lowers end-to-end cost from 13.31 to 10.31.
These results show that YAMATO not only preserves compressed state, but also prevents integrity errors from propagating into repeated actions and additional interaction cost.

Besides, Appendix~\ref{app:extended_experiments} provides extensive empirical analysis of DMC-5, covering robustness across runs and hyperparameters, YAMATO integrity and overhead, task difficulty, pricing and latency trade-offs, as well as representative success and failure trajectories.

\begin{table}[t]
\centering
\vspace{-10pt}
\caption{
Analysis of YAMATO over quality and integrity analysis on AppWorld.
}
\label{tab:yamato_main}
\vspace{-10pt}
\scriptsize
\setlength{\tabcolsep}{2.2pt}
\renewcommand{\arraystretch}{1.06}

\aboverulesep=0pt
\belowrulesep=0pt
\extrarowheight=0.2ex

\resizebox{\columnwidth}{!}{
\begin{tabular}{
l
>{\columncolor{gray!15}}l
l
>{\columncolor{purple!8}}l
l
>{\columncolor{blue!8}}l
}
\toprule
Metric
& \multicolumn{1}{>{\columncolor{gray!12}}c}{\makecell{No\\Compression}}
& \multicolumn{1}{c}{\makecell{ACON\\\citep{kang2025acon}}}
& \multicolumn{1}{>{\columncolor{purple!8}}c}{\makecell{ACON\\w/ YAMATO}}
& \multicolumn{1}{c}{\makecell{DMC-5\\w/o YAMATO}}
& \multicolumn{1}{>{\columncolor{blue!8}}c}{\makecell{DMC-5\\w/ YAMATO}} \\
\midrule

% \multicolumn{6}{l}{\textit{Overall outcome}} \\
\textit{Overall outcome} &&&&&\\

Task Accuracy $\uparrow$
& 77.0\%
& 67.3\%
& 72.0\%
& 69.1\%
& \textbf{78.2\%} \\

End-to-End Cost $\downarrow$
& 13.71
& 11.57
& 12.11
& 13.31
& \textbf{10.31} \\

Cost / Success $\downarrow$
& 0.1051
& 0.1024
& 0.1001
& 0.1147
& \textbf{0.0781} \\

% Latency $\downarrow$
% & \textbf{69s}
% & 78s
% & 80s
% & 100s
% & 104s \\

\midrule

\textit{Integrity of compression}&&&&&\\ 

Compression with Errors $\downarrow$
& --
& 4.2\% {\tiny(30/711)}
& 1.0\% {\tiny(9/894)}
& 3.0\% {\tiny(77/2581)}
& \textbf{0.04\%} {\tiny(1/2,460)} \\

Corrupted Execution-Critical Items $\downarrow$
& --
& 1.23\% {\tiny(48/3,893)}
& 0.35\% {\tiny(16/4,607)}
& 0.33\% {\tiny(82/24,771)}
& \textbf{0.01\%} {\tiny(3/22,551)} \\

Recall of Execution-Critical Items $\uparrow$
& --
& 63.3\% {\tiny(279/441)}
& 87.1\% {\tiny(608/698)}
& 77.7\% {\tiny(839/1079)}
& 98.3\% {\tiny(1008/1025)} \\

\midrule

\textit{Behavior of repeated failed actions}&&&&&\\

Runs with Repeat $\geq$5 $\downarrow$
& 1.2\% 
& 1.6\%
& 0.6\% 
& 1.6\% 
& \textbf{0.2\%} \\

Max Same-Action Repeats $\downarrow$
& 9
& 39
& 12
& 29
& \textbf{6} \\

\bottomrule
\end{tabular}
}

\vspace{1mm}
\scriptsize
\scriptsize
\textit{Note.} 
\textit{Corrupted items} are execution-critical items in the Decision View whose values disagree with the corresponding historical records.
\textit{Recall} is the fraction of correct, action-relevant execution-critical items that remain available in the Decision View.
\textit{Repeated failures} are subsequent identical or highly similar actions following an explicit execution error.
\vspace{-10pt}
\end{table}

\vspace{-10pt}

\subsection{Generalization to Agent Endpoints with Cached-Input Discounts}
\label{sec:cache_enabled_agent}

To evaluate DMC-5's generalization ability under deployments where the acting agent also benefits from prefix caching, we repeat the same AppWorld tasks using a Qwen3.8-27B base endpoint with discounted cached-input pricing. 
The results show that DMC-5 remains cost-effective even when the uncompressed agent already achieves a high cache-hit rate. 
At the actual price ratio of 20$\%$, the baseline achieves a 70.5$\%$ agent cache-hit rate, while DMC-5 achieves 51.5$\%$. 
Despite the lower agent-side cache reuse, DMC-5 reduces total cost from 35.72 to 32.38 (9.4$\%$) and increases task success from 71.5$\%$ to 73.9$\%$.

We further quantify the break-even boundary by varying the agent-side cached-input price while keeping the uncached-input price and observed token traces fixed. 
As shown in Table~\ref{tab:agent_cache_generalization}, the two methods reach equal cost at a cached-to-uncached price ratio of 9.1$\%$. 
Thus, DMC-5 remains cheaper as long as the cached-input price exceeds 9.1$\%$ of the uncached-input price. 
Under the actual pricing ratio of 20$\%$, DMC-5 operates comfortably within this cost-saving region. 

Overall, although DMC-5 is primarily designed for specialized agents without cache discounts, our results demonstrate that it also generalizes effectively to cache-enabled agent deployments.

\begin{table}[t]
\centering
\scriptsize
\setlength{\tabcolsep}{4.5pt}
\caption{DMC-5 with a cache-enabled agent Qwen3.8-27B on AppWorld. 
Cached Price Ratio denotes $p_m^{\mathrm{hit}}/p_m^{\mathrm{miss}}$.
The break-even analysis fixes the agent uncached-input price $p_m^{\mathrm{miss}}$ and uses the observed token traces.}
\label{tab:agent_cache_generalization}
\vspace{-10pt}
\resizebox{\linewidth}{!}{
\begin{tabular}{lcccccc}
\toprule
Setting &
Agent Cache Hit &
Cached Price Ratio &
Agent Cost &
Comp. Cost &
Total Cost &
Success \\
\midrule

\multicolumn{7}{l}{\textit{Observed operating point}} \\
No Compression
& \cellcolor{gray!15}70.5\%
& \cellcolor{gray!15}20.0\%
& \cellcolor{gray!15}35.72
& \cellcolor{gray!15}--
& \cellcolor{gray!15}35.72
& \cellcolor{gray!15}71.5\% \\

DMC-5
& \cellcolor{blue!8}51.5\%
& \cellcolor{blue!8}20.0\%
& \cellcolor{blue!8}29.49
& \cellcolor{blue!8}2.89
& \cellcolor{blue!8}\textbf{32.38} (\textbf{-9.4\%})
& \cellcolor{blue!8}\textbf{73.9\%} \\

\midrule
\multicolumn{7}{l}{\textit{Break-even for agent cached price ratio}} \\
No Compression
& 70.5\%
& \cellcolor{blue!8}9.1\%
& 29.80
& --
& 29.80
& 71.5\% \\

DMC-5
& 51.5\%
& \cellcolor{blue!8}9.1\%
& 26.91
& 2.89
& 29.80
& \textbf{73.9\%} \\

\bottomrule
\end{tabular}
}
\end{table}
\vspace{-5pt}

\subsection{Generalization Beyond Structured Schema}
\label{sec:tool_generalization}

DMC-5 operates on raw action, observation, and work-product records rather than depending on a predefined API schema. 
To evaluate its generality beyond structured function-calling settings,
we further test DMC-5 on PinchBench, which contains coding-style commands and file operations, and $\tau^2$-bench, which involves heterogeneous multi-turn tool interactions. 
Since DMC-5 only affects execution after compression is triggered, we report the subsets of tasks that reach the compression threshold, thereby isolating the effect of context compression.

\begin{table}[t]
\centering
\scriptsize
\caption{Generalization to coding-style and heterogeneous tool-use environments.
Results are reported on compression-triggered subsets, containing 40 PinchBench
tasks and 64 $\tau^2$-bench tasks.}
\label{tab:tool_generalization}
\resizebox{\linewidth}{!}{
\begin{tabular}{llccccc}
\toprule
Benchmark &
Setting &
Agent Input &
Agent Cost &
Comp. Cost &
Total Cost &
Successful Tasks \\
\midrule
\multirow{2}{*}{PinchBench}
& No Compression
& 13.14M
& 4.39
& --
& 4.39
& 37/40 \\
& \cellcolor{blue!8}DMC-5
& \cellcolor{blue!8}\textbf{10.51M} (\textbf{-20.0\%})
& \cellcolor{blue!8}3.55
& \cellcolor{blue!8}0.48
& \cellcolor{blue!8}\textbf{4.03} (\textbf{-8.2\%})
& \cellcolor{blue!8}\textbf{37/40} \\
\midrule
\multirow{2}{*}{$\tau^2$-bench}
& No Compression
& 10.59M
& 3.66
& --
& 3.66
& 42/64 \\
& \cellcolor{blue!8}DMC-5
& \cellcolor{blue!8}\textbf{8.27M} (\textbf{-21.9\%})
& \cellcolor{blue!8}2.85
& \cellcolor{blue!8}0.50
& \cellcolor{blue!8}\textbf{3.35} (\textbf{-8.5\%})
& \cellcolor{blue!8}\textbf{43/64} \\
\bottomrule
\end{tabular}}
\vspace{-12pt}

\end{table}

DMC-5 reduces agent input by 20.0$\%$ and 21.9$\%$, and total cost by 8.2$\%$ and 8.5$\%$ on PinchBench and $\tau^2$-bench, respectively, while preserving or slightly improving task success. 
The smaller savings reflect shorter trajectories with fewer compression opportunities, consistent with the trajectory-length analysis in Table~\ref{tab:compression_count}.

Overall, these results indicate that DMC-5 generalizes beyond structured API-based agents, and that its efficiency gains depend more strongly on trajectory length and repeated history processing than on the specific form of the tool interface.
\vspace{-5pt}

\section{Related Work}
\label{sec:related_work}
\vspace{-5pt}

\textbf{Agent context compression and memory.}
Prompt compression shortens an LLM's input by token selection or summaries. 
LLMLingua and LongLLMLingua use smaller models to identify useful content, while RECOMP compresses retrieved evidence~\citep{jiang2023llmlingua,jiang2023longllmlingua,xu2023recomp}. 
Agent trajectories require repeated compression because they grow after every action. MemGPT manages context as tiered memory, and HiAgent groups interactions by subgoal~\citep{packer2023memgpt,hu2024hiagent}. 
ACON optimizes compression guidelines for histories and observations and distills them into smaller compressors; AgentDiet removes useless, repeated, or expired content; AdaCoM trains an external context manager; and Focus lets an agent trigger consolidation~\citep{kang2025acon,xiao2026agentdiet,yi2026adacom,verma2026focus}. 
MemAgent streams long inputs through a fixed-size memory trained with reinforcement learning~\citep{yu2025memagent}.
These methods shorten the acting model's context by selecting, rewriting, or externally managing prior state.
Instead, DMC-5 keeps the raw trajectory append-only for repeated full-history integration, preserving a stable cached prefix for the compressor while providing the specialized agent with a separate compact Decision View.

\textbf{Cache-aware LLM inference.}
Serving systems reduce repeated-prefix cost by managing KV caches. KVFlow schedules cache retention for agent workflows, CacheBlend reuses non-prefix caches through selective recomputation, and ThinKV compresses reasoning-model caches with adaptive quantization and eviction~\citep{pan2025kvflow,yao2024cacheblend,ramachandran2025thinkv}. 
LRAgent supports cache sharing across LoRA-specialized agents by separating shared and adapter-dependent states~\citep{jeon2026lragent}. 
CAPC is particularly relevant to our setting, as it shows that query-dependent prompt compression can disrupt prefix reuse and explicitly optimizes compression for cache compatibility~\citep{song2026capc}.
However, CAPC focuses on cache-aware compression within individual prompts, whereas DMC-5 considers repeated trajectory compression across agent turns.
By maintaining an append-only full-history view, DMC-5 enables trajectory-level history integration to continuously reuse existing prefix caches and benefit from serving-side cache optimizations.

\textbf{Multi-model collaboration and agent efficiency.}
Model cascades and routers reduce cost by assigning requests to models with different capabilities. 
FrugalGPT learns API cascades, and RouteLLM selects between stronger and weaker models from preference data~\citep{chen2023frugalgpt,ong2024routellm}. 
DUET uses one model to produce a short reasoning signal that another turns into an answer~\citep{chen2026duet}. 
Research on agent evaluation also argues that quality and resource use should be measured together~\citep{kapoor2024agents}. 
Unlike routing or cascade methods that decide which model should answer a request, DMC-5 keeps both models active with fixed asymmetric roles.
The general model integrates full history into a compact Decision View for specialized model to generate action, reducing end-to-end cost while keeping task performance of long-horizon agents.

\vspace{-5pt}
\section{Conclusion}
\label{sec:conclusion}
\vspace{-5pt}

We presented \textbf{DMC-5} (Dual-Model Collaboration with CaChe-Aware Context Compression), a dual-model framework for reducing end-to-end cost of long-horizon agents by assigning full-history integration to general compressor and action generation to specialized agent.
DMC-5 uses Append-only Canonical View to preserve full-history access and prefix-cache reuse, compact Decision View to reduce repeated agent input, and \textbf{YAMATO} (integritY-Aware Memory Abstractions for Tool Operations) to protect execution-critical information during lossy compression.
Across AppWorld, OfficeBench, and 8-objective QA, DMC-5 reduces agent input by 38.5$\sim$57.0$\%$ and end-to-end cost by 14.5$\sim$28.0$\%$, while improving task performance by 1.0$\sim$5.8$\%$ over uncompressed baseline.

\subsection*{AI use statement}

Generative AI tools were used to assist with language polishing, code development,
and visualization.
The research ideas, methodology, experimental design, and
scientific claims were developed by the human authors, and all cited references
were manually read and selected by the authors.

AI-assisted code was written under human direction and subsequently reviewed and
verified by the authors. Experimental results were collected and aggregated using
deterministic scripts rather than generative AI. Figures were generated from
verified experimental data using plotting scripts partially assisted by AI tools,
and all reported values and visualizations were manually checked.

All AI-assisted text and code were reviewed by the authors, who take full
responsibility for the final content of this work.

\subsection*{Ethics statement}

This work uses established research benchmarks and does not involve human subjects,
private or sensitive data, or other specific ethical concerns.

\subsection*{Reproducibility Statement.}
We provide the implementation of DMC-5 and executable evaluation script for AppWorld in the supplementary materials submitted through OpenReview.
Detailed benchmark settings, model configurations, prompts, compression hyperparameters, pricing coefficients, and evaluation protocols are documented in Appendix~\ref{app:benchmark_protocols}. 
These materials are intended to enable reproduction of the main experimental results.

\providecommand{\venueICLR}{ICLR}
\providecommand{\venueICML}{ICML}
\providecommand{\venueNeurIPS}{NeurIPS}
\providecommand{\venueACL}{ACL}
\providecommand{\venueEMNLP}{EMNLP}
\providecommand{\venueFindingsACL}{Findings of ACL}
\providecommand{\venueEuroSys}{EuroSys}
\providecommand{\venueMLSys}{MLSys}
\providecommand{\venueUIST}{UIST}
\providecommand{\venueSOSP}{SOSP}
\providecommand{\venueSIGCOMM}{ACM SIGCOMM}

\providecommand{\journalNeuralComp}{Neural Comput.}
\providecommand{\journalTMLR}{TMLR}
\providecommand{\journalTACL}{TACL}
\providecommand{\journalPACMSE}{Proc. ACM Softw. Eng.}
\bibliography{iclr2027_conference}
\bibliographystyle{iclr2027_conference}

\newpage

\section{Appendix}
\label{sec:appendix}

This appendix specifies the evaluation protocol, implementation, and accounting
used in DMC-5. It also reports uncertainty estimates, mechanism-level ablations,
efficiency analyses, and representative trajectory cases.

\subsection{YAMATO Implementation Details}
\label{app:yamato_details}

YAMATO augments lossy semantic compression with execution-critical information
that should not be reconstructed from model-generated text.
It introduces no additional LLM calls: all model-generated fields are produced
within the existing compressor call, while pointer restoration, work-product
retention, failure tracking, and provenance checks are implemented
deterministically in code.

At compression round $r$, the compressor $V$ reads the task instruction and the
eligible prefix of the Append-only Canonical View, and outputs
\begin{equation}
    \widehat{M}_r = (Z_r,\Pi_r),
    \label{eq:app_yamato_output}
\end{equation}
where $Z_r$ is a structured semantic summary and $\Pi_r$ contains references
to canonical records that should be restored exactly.
YAMATO organizes the compressed state through three complementary mechanisms:
(i) a structured summary with turn provenance,
(ii) exact restoration of selected canonical records, and
(iii) a persistent failure ledger.
The resulting compact state is then combined with recent uncompressed turns to
form the Decision View.

\subsubsection{Canonical Records and Preservation Candidates}
\label{app:yamato_candidates}

Each canonical turn $\xi_i=(a_i,o_i)$ contains the agent action $a_i$ and its
observation $o_i$, both of which remain unchanged in the Append-only Canonical
View.
YAMATO maintains a turn registry that maps each turn index $i$ to its raw action
and observation strings.

Specifically, each new action--observation turn is serialized once when it is
appended to the Canonical View.
Deterministic rules add a type marker to the turn header:
\texttt{(API DOC)} and \texttt{(FILE CONTENT)} denote literal tool outputs,
\texttt{(WORK PRODUCT)} denotes artifact-producing or artifact-updating actions,
and \texttt{(FAILED)} denotes unsuccessful operations.
Once appended, earlier turns are neither regenerated nor relabeled, preserving
the byte-identical historical prefix required for cache reuse.

For example, an API-documentation turn may be serialized as:
\begin{verbatim}
--- Turn 7 (API DOC) ---
[ASSISTANT ACTION]
get_api_documentation("calendar.create_event")

[TOOL RESULT]
create_event(calendar_id, title, start_time)
\end{verbatim}

These markers guide the compressor in identifying candidate records for exact
preservation.
When a marked record is needed in the Decision View, the compressor can
reference it with a turn pointer instead of copying the full raw content into
the semantic summary.
For example, it may emit \texttt{<<TURN\_7\_RESULT>>}, which YAMATO resolves
deterministically to the original observation stored in the Canonical View.

\subsubsection{Structured Summary with Turn Provenance}
\label{app:yamato_summary}

The semantic summary adopts an ACON-style~\citep{kang2025acon} structured history format, with its fields adapted to support DMC-5's provenance tracking, exact-state restoration, and persistent failure state.
Together with the Append-only Canonical View, it forms the semantic component of DMC-5's state representation.

In particular, the semantic component $Z_r$ compresses the task state into five fixed fields:
\textsc{Reasoning}, \textsc{Completed}, \textsc{Pending},
\textsc{Failed Attempts}, and \textsc{Key Data}.
They respectively record the current task state, completed progress, remaining
steps, previous unsuccessful attempts, and compact information that may be
needed by later actions.
A simplified format is shown below:

\begin{verbatim}
### REASONING
- ...

### COMPLETED
- ...

### PENDING
- ...

### FAILED ATTEMPTS
- ...

### KEY DATA
- ...
\end{verbatim}

To distinguish environment-grounded information from agent-side derivations,
the compressor attaches turn-level provenance tags to summarized claims.
Table~\ref{tab:app_yamato_provenance} summarizes the tags.

\begin{table}[h]
\centering
\small
\caption{Turn-provenance tags used in the structured YAMATO summary.}
\label{tab:app_yamato_provenance}
\setlength{\tabcolsep}{5pt}
\renewcommand{\arraystretch}{1.10}
\begin{tabular}{@{}p{0.20\columnwidth}
                p{0.28\columnwidth}
                p{0.43\columnwidth}@{}}
\toprule
Tag & Source & Meaning \\
\midrule

\texttt{[O@i]}
& Observation $o_i$
& The summarized information is grounded in a tool or environment observation
at turn $i$. \\

\texttt{[D@i]}
& Agent-side state at turn $i$
& The information is an agent-derived plan, inference, or intermediate
conclusion associated with turn $i$, rather than a directly observed result. \\

\texttt{[UNSOURCED]}
& No explicit turn reference
& A \textsc{Key Data} entry has no recognized source reference in the
compressed summary. \\

\bottomrule
\end{tabular}
\end{table}

The \texttt{[O@i]} and \texttt{[D@i]} tags are generated as part of the same
compressor output.
The compressor is instructed to preserve the distinction between information
observed from the environment and information derived by the agent when
rewriting historical state.

For \textsc{Key Data}, YAMATO additionally applies a lightweight deterministic
format check.
If an entry contains no recognized turn-level source reference, the code marks
it as \texttt{[UNSOURCED]}.
This check does not determine whether the claim is factually correct and does
not modify an existing \texttt{[O@i]} or \texttt{[D@i]} tag.
Its purpose is only to make missing provenance explicit to the acting agent.

We denote the resulting provenance-aware semantic summary as
\begin{equation}
    \overline{Z}_r
    =
    \operatorname{Prov}(Z_r),
    \label{eq:app_yamato_provenance}
\end{equation}
where $\operatorname{Prov}(\cdot)$ denotes the deterministic provenance-format
check described above.

Turn provenance complements exact raw-record restoration:
$\overline{Z}_r$ preserves compact semantic state together with its historical
source, while pinned pointers are used when execution-critical content should
be copied directly from the Canonical View.

\subsubsection{Pinned Records and Work-Product Restoration}
\label{app:yamato_pinned}

YAMATO protects three types of execution-critical information:
\emph{tool observations}, \emph{agent actions}, and \emph{work products}.
For these records, the compressor emits short pointers to the corresponding
canonical turns rather than reproducing their contents in the semantic summary.
YAMATO then resolves the pointers deterministically.

Let
\begin{equation}
    \Pi_r
    =
    \Pi_r^{\mathrm{act}}
    \cup
    \Pi_r^{\mathrm{obs}}
    \cup
    \Pi_r^{\mathrm{work}}
\end{equation}
denote the action, observation, and work-product pointers emitted at compression
round $r$.
Table~\ref{tab:app_yamato_pointer_schema} summarizes their semantics.

\begin{table*}[t]
\centering
\small
\caption{YAMATO pointer interfaces for restoring execution-critical state from
the Append-only Canonical View.}
\label{tab:app_yamato_pointer_schema}
\setlength{\tabcolsep}{5pt}
\renewcommand{\arraystretch}{1.12}
\begin{tabular}{@{}p{0.22\textwidth}
                p{0.20\textwidth}
                p{0.28\textwidth}
                p{0.22\textwidth}@{}}
\toprule
Pointer & Canonical target & Restored content & Handling \\
\midrule

\texttt{<<TURN\_i\_RESULT>>}
& Observation $o_i$
& Exact tool or environment output, such as API responses, logs,
documentation, returned text, or file contents.
& Direct lookup and restoration. \\

\texttt{<<TURN\_i\_ACTION>>}
& Action $a_i$
& Exact tool call, command, arguments, or write/edit operation.
& Direct lookup and restoration. \\

\texttt{<<TURN\_i\_WORK>>}
& Work-product action at turn $i$
& Exact artifact-related operation associated with a file.
& Lookup followed by code-side matching and ordered aggregation of operations
on the same file. \\

\bottomrule
\end{tabular}
\end{table*}

For action and observation pointers, YAMATO directly retrieves the referenced
raw records:
\begin{align}
    E_r^{\mathrm{act}}
    &=
    \mathcal{R}
    \left(
        \Pi_r^{\mathrm{act}};
        \tau_{b_r}
    \right), \\
    E_r^{\mathrm{obs}}
    &=
    \mathcal{R}
    \left(
        \Pi_r^{\mathrm{obs}};
        \tau_{b_r}
    \right),
    \label{eq:app_yamato_direct_restore}
\end{align}
where $\mathcal{R}$ denotes deterministic lookup from the Canonical View and
$b_r$ is the last canonical turn included in compression round $r$.
The referenced action or observation is copied verbatim into the Decision View.

Work-product pointers use the same lookup mechanism, but require one additional
step because an artifact may evolve across multiple turns.
YAMATO first resolves the referenced work-product records:
\begin{equation}
    \widetilde{E}_r^{\mathrm{work}}
    =
    \mathcal{R}
    \left(
        \Pi_r^{\mathrm{work}};
        \tau_{b_r}
    \right).
    \label{eq:app_yamato_work_lookup}
\end{equation}

For each resolved work-product record, YAMATO identifies its artifact path and
matches other work-product operations on the same file in the Canonical View.
These operations are then concatenated in their original execution order:
\begin{equation}
    E_r^{\mathrm{work}}
    =
    \mathcal{G}
    \left(
        \widetilde{E}_r^{\mathrm{work}};
        \tau_{b_r}
    \right),
    \label{eq:app_yamato_work_aggregate}
\end{equation}
where $\mathcal{G}$ denotes deterministic path matching and ordering over the
eligible Canonical View.
This preserves the sequence of writes and edits that defines the current work
product without asking the compressor to reconstruct the artifact from text.

The restored execution-critical context is therefore
\begin{equation}
    E_r
    =
    E_r^{\mathrm{act}}
    \oplus
    E_r^{\mathrm{obs}}
    \oplus
    E_r^{\mathrm{work}}.
    \label{eq:app_yamato_exact_state}
\end{equation}

\subsubsection{Persistent Failure Ledger}
\label{app:yamato_failure_ledger}

Lossy compression may discard previous failed attempts, causing the agent to
repeat the same ineffective action.
YAMATO therefore maintains a persistent failure ledger $L_r$ across compression
rounds:
\begin{equation}
    L_r
    =
    \operatorname{Tail}_{N}
    \left(
        L_{r-1}
        \oplus
        L_r^{\mathrm{new}}
    \right),
    \label{eq:app_yamato_ledger}
\end{equation}
where $L_r^{\mathrm{new}}$ contains failures observed since the previous
compression round and $\operatorname{Tail}_{N}(\cdot)$ retains only the most
recent $N$ entries.
We set $N=40$ by default.
The ledger serves as persistent context for avoiding repeated mistakes, but does
not prevent the agent from retrying an operation when later context suggests a
corrected action.

\subsubsection{Decision-View Construction}
\label{app:yamato_decision_view}

After provenance processing, exact-record restoration, and failure-state update,
YAMATO assembles the compact state at compression round $r$ as
\begin{equation}
    C_r
    =
    \overline{Z}_r
    \oplus
    E_r
    \oplus
    L_r,
    \label{eq:app_yamato_core}
\end{equation}
where $\overline{Z}_r$ is the provenance-aware semantic summary,
$E_r$ contains the restored execution-critical records, and $L_r$ is the
persistent failure ledger.

Suppose compression round $r$ is triggered at agent turn $t_r$.
The most recent $K$ turns are excluded from the compressed prefix, so
\begin{equation}
    b_r = t_r-K,
\end{equation}
where $b_r$ is the last canonical turn integrated into the compressed state.
Let $r(t)$ denote the latest completed compression round at agent turn $t$.
The Decision View is then
\begin{align}
    M_t
    &=
    C_{r(t)}
    \oplus
    \tau^{\mathrm{raw}}_{b_{r(t)}+1:t},
    \\
    X_t^U
    &=
    s_U
    \oplus q
    \oplus M_t,
    \label{eq:app_yamato_view}
\end{align}
where $\tau^{\mathrm{raw}}_{b_{r(t)}+1:t}$ contains the recent raw turns that
were excluded from compression together with any turns generated afterward.

Thus, $C_r$ remains fixed between two compression rounds, while new actions and
observations are appended unchanged to the raw suffix until the next refresh.
The resulting Decision View combines four complementary information sources:
(i) a compact semantic summary for task progress and long-range state,
(ii) exact restored records for execution-critical information,
(iii) a persistent ledger of previous failures, and
(iv) recent raw interactions.
YAMATO therefore does not make the complete trajectory lossless; instead, it
uses semantic compression for ordinary history while protecting information
whose corruption or omission may affect subsequent tool execution.

\textbf{Illustrative example.}
Listing~\ref{lst:decision_view_case} illustrates the Decision View for an
OfficeBench task.
Assume that compression round $r$ is triggered after turn $9$ with $K=1$.
The compressed prefix therefore covers turns $1$--$8$ ($b_r=8$), while turn
$9$ remains in the recent raw suffix.
Lines beginning with \texttt{\%\%} are annotations for illustration and are
not visible to the agent.

\begin{lstlisting}[
  caption={Illustrative Decision View for an OfficeBench task.},
  label={lst:decision_view_case},
  basicstyle=\ttfamily\scriptsize,
  frame=single,
  breaklines=true,
  columns=fullflexible,
  keepspaces=true,
  showstringspaces=false
]
%% C_r = Zbar_r + E_r + L_r

<HISTORY_SUMMARY>

### COMPLETED
- Located the Q2 revenue value [O@2]
- Created Q2_Sales_Report.docx [O@8]

### PENDING
- Export the report document to PDF [D@8]

### FAILED ATTEMPTS
- T4: write_file failed because "path" was missing

### KEY DATA
- net revenue: 182400.00 [O@2]
- report path:
  /workspace/Q2_Sales_Report.docx [O@8]

### PINNED RECORD
%% Restored from <<TURN_2_RESULT>>

File: quarterly_sales.xlsx
Sheet: Q2
A6: Net Revenue
B6: 182400.00

### WORK PRODUCT
%% Restored from the canonical work-product history

{"action":"write_file",
 "path":"/workspace/Q2_Sales_Report.docx",
 "content":"Q2 Sales Report\nNet Revenue: 182400.00"}

</HISTORY_SUMMARY>


%% tau_raw_{9:9}

[ASSISTANT ACTION, T9]
{"action":"convert_to_pdf",
 "input_path":"/workspace/Q2_Sales_Report.docx",
 "output_path":"/workspace/Q2_Sales_Report.pdf"}

[TOOL RESULT, T9]
Conversion completed:
/workspace/Q2_Sales_Report.pdf
\end{lstlisting}

The upper block corresponds to $C_r$.
Its semantic fields come from $\overline{Z}_r$, the spreadsheet record and
work-product state are restored through $E_r$, and the failed
\texttt{write\_file} attempt is retained in $L_r$.
For example, \texttt{[O@2]} marks information grounded in the tool observation
at turn 2, whereas \texttt{[D@8]} denotes an agent-derived pending decision.

Turn $9$ is excluded from the compressed prefix as the recent raw turn and
therefore belongs to $\tau^{\mathrm{raw}}_{9:9}$ rather than $C_r$.
The agent receives both parts in the same Decision View.
As further turns are generated, they are appended to this raw suffix until the
next compression round constructs a new compact state.

\subsection{Evaluation Details}
\label{app:benchmark_protocols}

% Required packages:
% \usepackage{booktabs,adjustbox,makecell}
% \usepackage{amsmath}

\subsubsection{Benchmark and task configurations.}
Table~\ref{tab:app_benchmark_protocols} summarizes our evaluation sets.
We evaluate AppWorld on all 168 tasks in \texttt{test\_normal\_split}, OfficeBench on 192 subtask instances from the combined training and testing partitions, and 8-objective QA on all 200 tasks from its training and test partitions.
DMC-5 requires no model parameter updates.
We tune only the compression prompt on the AppWorld training split, then freeze it and apply the same prompt unchanged to all three benchmarks.
Since no training is performed on OfficeBench or 8-objective QA, we combine their available training and test splits to maximize evaluation coverage.
For a controlled comparison, all methods use the same task sets and a uniform budget of at most 50 agent interaction steps per task.
All reported results are averaged over three independent runs.

\begin{table}[h]
\centering
\small
\caption{Benchmark overview.}
\label{tab:app_benchmark_protocols}
\setlength{\tabcolsep}{4pt}
\renewcommand{\arraystretch}{1.15}
\begin{adjustbox}{max width=\textwidth}
\begin{tabular}{llrcrl}
\toprule
Benchmark
& Evaluated split
& \#Tasks
& Task grouping
& \makecell{Max\\steps}
& Metric \\
\midrule
AppWorld
& test\_normal\_split
& 168
& \makecell{57 / 48 / 63\\Easy / Medium / Hard}
& 50
& Accuracy \\
OfficeBench
& Training $+$ testing
& 192
& \makecell{86 / 51 / 55\\1-app / 2-app / 3-app}
& 50
& Accuracy \\
8-objective QA
& Train $+$ test
& 200
& --
& 50
& EM / F1 \\
\bottomrule
\end{tabular}
\end{adjustbox}
\end{table}
\paragraph{AppWorld}~\citep{trivedi2024appworld} provides a stateful execution
environment spanning nine everyday applications, including Spotify,
Gmail, and SimpleNote, with 457 APIs and approximately 100 simulated
users~\citep{kang2025acon}.
An agent completes user requests through executable actions that
retrieve information and modify application state, often requiring
coordination across applications.
Our evaluation uses 168 \texttt{test\_normal\_split} tasks:
57 easy, 48 medium, and 63 hard.
Correctness is measured using the benchmark-defined task goal
completion score, reported as accuracy (\%).
This benchmark tests whether compressed context preserves the
information and execution state needed to complete multi-step
application workflows.

\paragraph{OfficeBench}~\citep{wang2024officebench} evaluates office automation
workflows involving applications such as Word, Excel, PDF, Calendar,
and Email, together with utilities such as Shell and Calculator.
Tasks are grouped into 86 1-app, 51 2-app, and 55 3-app categories according
to the number of applications involved.
We evaluate 192 subtask instances from the combined training and
testing partitions used in our implementation.
Correctness is assessed using the benchmark-provided accuracy
functions, which evaluate the outcomes of the requested office
operations.
We report accuracy (\%) aggregated over the evaluated instances.
These workflows test whether the agent retains relevant document
content, intermediate results, and task progress across successive
tool interactions.

\paragraph{8-objective QA}~\citep{zhou2025mem1, kang2025acon}
groups eight distinct questions from
NaturalQuestions into a single task.
The agent interacts with a search tool and must return a consolidated
set of answers to all eight questions.
Unlike single-question multi-hop QA, this setting requires tracking
multiple question--answer pairs and unresolved objectives within
the same interaction history.
We evaluate all 100 training tasks and 100 test tasks, giving
200 composite tasks and 1,600 question instances.
Long context accumulates through successive search queries,
retrieved evidence, and intermediate reasoning, while answers
obtained earlier must remain available for the final response.
Following~\citet{kang2025acon}, we report exact match (EM) and F1
separately, averaging each metric over the eight questions in a task
and then over all evaluated tasks.

\begin{table}[h]
\centering
\small
\caption{Representative tasks from the evaluated benchmarks.}
\label{tab:app_benchmark_examples}
\setlength{\tabcolsep}{5pt}
\renewcommand{\arraystretch}{1.15}

\begin{tabularx}{\textwidth}{
    @{}ll>{\raggedright\arraybackslash}X@{}
}
\toprule
Benchmark & Category & Example Task \\
\midrule

\multirow{3}{*}{AppWorld}
& Easy
& \rule{0pt}{3.5ex}Remove all the songs that I have liked from my Spotify queue, and then start the player. \\

& Medium
& \rule{0pt}{3.5ex}Reset friends on venmo to be the same as my friends in my phone. Befriend and unfriend as needed. \\

& Hard
& \rule{0pt}{3.5ex}My roommates and I went for a dinner yesterday. I paid for the taxi back and forth (total \$60) and Nancy paid for everyone's food (total \$128). Both food and commute are supposed to be shared equally among all. Make necessary payment requests with a note "For Taxi", and a payment to Nancy with a note "For Food", on venmo. \\

\midrule

\multirow{3}{*}{OfficeBench}
& 1-app
& \rule{0pt}{3.5ex} Add a meeting to Bob's calendar at 5/17/2024 10:30 a.m to 11:00 a.m. \\
% TODO: Insert one evaluated 1-app task.

& 2-app
& \rule{0pt}{3.5ex}Schedule a team training session for all participants from excel file, create calendar events for each member in their name. \\
% TODO: Insert one evaluated 2-app task.

& 3-app
& \rule{0pt}{3.5ex}Convert Excel spreadsheet containing project tasks to task.pdf and email Alice with contents from the pdf file for review. \\
% TODO: Insert one evaluated 3-app task.

\midrule

8-objective QA
& Multi-question
& \rule{0pt}{5ex}What category was hurricane charley when it hit florida?; where in the bible is the book of revelation?; who played skeletor in the movie masters of the universe?; who has the most votes for the all star game?; when was the first mad max movie release?; the cast of don't tell mom the babysitter's dead?; what does the adrenal gland produce that is necessary for the sympathetic nervous system to function?; who played ice queen in chronicles of narnia? \\
% TODO: Insert one evaluated composite task with all eight questions.

\bottomrule
\end{tabularx}
\end{table}

\subsubsection{Models, Endpoints, and Pricing}
\label{app:model_pricing}

All models are accessed through API endpoints on Alibaba Cloud Model Studio
(Bailian).\footnote{\url{https://www.alibabacloud.com/en}}
The specialized agent is based on Qwen3.5-27B and deployed as a LoRA-adapted
endpoint after fine-tuning on private agent data.
We verified that no evaluation instances from the three benchmarks are included in the fine-tuning data.
DeepSeek-V4-flash serves as the default general compressor, while
Qwen3.7-flash is used in the model-role ablation.
Unless otherwise stated, the same specialized agent is used across all
compression methods.

Table~\ref{tab:app_models_pricing} summarizes the pricing used for end-to-end cost accounting.
We use the Alibaba Cloud Model Studio (Bailian) pricing in effect at the time of evaluation.
Model prices may change over time, and all reported costs are computed using this fixed pricing snapshot.

\begin{table}[h]
\centering
\small
\caption{Model roles and pricing used for end-to-end cost accounting.
Prices are reported per one million tokens and correspond to the pricing
in effect at the time of evaluation.}
\label{tab:app_models_pricing}
\setlength{\tabcolsep}{5pt}
\renewcommand{\arraystretch}{1.15}
\begin{adjustbox}{max width=\textwidth}
\begin{tabular}{llrrrl}
\toprule
Role
& Model
& \makecell{Uncached\\input}
& \makecell{Cached\\input}
& Output
& Deployment \\
\midrule
Specialized agent
& Qwen3.5-27B (LoRA)
& 0.30
& 0.30
& 2.40
& Fine-tuned endpoint \\

Primary compressor
& DeepSeek-V4-flash
& 0.50
& 0.10
& 1.00
& API endpoint \\

Alternative compressor
& Qwen3.7-flash
& 0.10
& 0.02
& 0.40
& API endpoint \\
\bottomrule
\end{tabular}
\end{adjustbox}
\end{table}

\subsubsection{Inference Configuration, Prompts, and Compression Settings}
\label{app:inference_config}

\paragraph{Inference configuration.}
All compressor calls use greedy decoding with temperature $0$ to minimize
stochastic variation in the compressed state.
We disable the model's internal thinking mode
(\texttt{enable\_thinking=false}), since trajectory compression is formulated
as structured information extraction and state preservation rather than
open-ended reasoning.
For the specialized agent, we retain the decoding configuration recommended
by the deployed API endpoint, including its default temperature and
benchmark-consistent thinking setting.
We do not fix generation seeds.
Instead, all reported results are obtained from three independent runs to
capture the stochasticity of API-based agent execution.

\paragraph{Hyperparameters.}
Unless otherwise specified, DMC-5 uses the same compression configuration
across all three benchmarks.
The most recent $K=1$ interaction turn is always retained verbatim in the
Decision View.
A new compression round is triggered when either
(i) three new interaction turns have accumulated since the previous
compression, or
(ii) the newly accumulated uncompressed trajectory exceeds $6{,}000$ tokens.
Thus, with compression interval $\Delta=3$ and token threshold
$T=6{,}000$, the trigger condition is
\begin{equation}
    \mathrm{Compress}(t)
    =
    \mathbb{I}
    \left[
        \Delta_{\mathrm{turn}}(t) \geq 3
        \;\lor\;
        \Delta_{\mathrm{token}}(t) \geq 6{,}000
    \right].
\end{equation}
At each compression round, the compressor reads the complete eligible prefix
from the Append-only Canonical View, while the most recent turn remains
uncompressed.

\begin{table}[h]
\centering
\small
\caption{Default hyperparameters.}
\label{tab:app_inference_config}
\setlength{\tabcolsep}{6pt}
\renewcommand{\arraystretch}{1.12}
\begin{tabular}{ll}
\toprule
Setting & Value \\
\midrule
Compressor temperature & $0$ (greedy) \\
Compressor thinking & Disabled \\
Recent raw turns $K$ & $1$ \\
Compression interval $\Delta$ & $3$ turns \\
Token trigger $T$ & $6{,}000$ new tokens \\
Agent decoding & Endpoint / benchmark default \\
\bottomrule
\end{tabular}
\end{table}

\paragraph{Prompts and input construction.}
The specialized agent retains the original benchmark-specific agent prompt
without modification.
Across benchmarks, DMC-5 follows the same compressor-prompt design and input
organization, while only the task- and tool-specific content varies.
The compressor summarizes task progress and pending work, preserves
execution-critical state and provenance, and emits raw-record pointers for
information that should remain byte-exact.
A representative compressor prompt is shown in
Listing~\ref{lst:compressor_prompt}.
\begin{lstlisting}[style=prompt,
caption={An example compressor prompt used for trajectory compression.},
label={lst:compressor_prompt}]
You are a context compression agent for a multi-step, tool-using AI agent.

You will receive a task instruction followed by the agent's full interaction
trajectory: a sequence of turns, each containing the agent's action
([ASSISTANT ACTION], a tool call like `tool({...arguments...})`) and the
corresponding tool result / observation ([TOOL RESULT]). Turns marked with
(FILE CONTENT) contain raw source material in their tool result: the exact
contents of a file read from disk, or the readable text of a fetched web
page. Turns marked with (WORK PRODUCT) contain file contents the agent
wrote or edited itself (e.g. an analysis script) -- the artifact the next
steps keep modifying. Turns marked with (FAILED) produced an error or an
empty result.

Produce a structured summary of the trajectory that preserves all critical
decision-making context for the agent's next step. You must NEVER add
information that was not present in the original trajectory.

PROVENANCE RULES -- you are a scribe, not a judge. The trust level of any
information may only be preserved or downgraded by summarisation, never
upgraded:
1. Tag statements with their source class (shorthand):
   - [O@i] -- observed directly from the tool result at turn i.
   - [D@i] -- a plan, rule, or conclusion the AGENT itself formed at turn i;
     UNVERIFIED, may be wrong; never present it as established fact.
   You yourself must NOT introduce any new conclusion.
2. NEVER restate task requirements in your own words -- write
   "per task instruction" instead of paraphrasing (paraphrase is how
   constraints get silently mutated).
3. Uncertainty must survive: keep hedges ("probably", "only page 0
   fetched"); never state anything more confidently than your source.

BREVITY RULES -- output tokens are expensive; every character must earn
its place:
4. Telegraphic note form. No full sentences, no articles, no pleasantries.
   e.g. "csv parsed, 250 rows, 2024 monthly [O@5]" not "I have successfully
   parsed the CSV file which contains 250 rows of monthly data."
5. Single home per value: exact values (totals, dates, names, URLs, IDs)
   live ONLY in KEY DATA. Other sections refer to them by name, never
   repeat them.
6. Do NOT list turns marked (FAILED) in FAILED ATTEMPTS -- they are
   recorded automatically by the system. Only list dead ends that are NOT
   marked (FAILED), e.g. a search that returned results which turned out
   irrelevant.

VERBATIM PRESERVATION RULES:
- Exact formats matter. Numbers, column names, dates, URLs, and quoted
  facts must NEVER be paraphrased or rounded. Copy them character-by-
  character, or use a placeholder.
- To preserve an entire tool result verbatim, do NOT copy its text.
  Instead output a placeholder on its own line: <<TURN_i_RESULT>>
  (i = turn number). It is auto-replaced with the original text --
  cheaper and always exact.
- In PINNED CONTENT, pin (via placeholder) every (FILE CONTENT) turn whose
  material is still needed for the remaining steps in PENDING -- especially
  files whose values must be cited in the final answer/report, and fetched
  pages whose facts have NOT yet been extracted into KEY DATA. Do NOT pin
  files already fully processed, nor pages whose relevant facts are fully
  captured in KEY DATA -- pinning processed sources only bloats the view.
- If the trajectory ends with a "=== CODE-CONSUMED SOURCES ===" note, the
  listed files have already been processed by code (a later script reads
  them from disk; subsequent steps work from script output, never from the
  raw text). Do NOT pin those turns -- their value now lives in the script
  outputs and in KEY DATA.
- (WORK PRODUCT) turns: the LATEST one is auto-pinned by the system -- do
  NOT pin it again. An OLDER work product still needed (e.g. a script that
  will be edited again while a newer one is also pinned) may be pinned via
  <<TURN_i_ACTION>> on its own line.
- When the pinned work product is an edit chain (a write plus subsequent
  edits to the same file), the system pins the WHOLE chain -- base write
  plus patches in order -- and appends it after your summary. That chain IS
  the current file state: never list "re-read the script/file" as a
  PENDING step and never claim the file content was lost.
- This is critical for FAILED shell/python turns: paraphrasing a failed
  command ("openpyxl import failed") does NOT tell the agent WHICH package
  or flag broke -- the agent will retry near-identical variants and burn
  steps. The exact previous command is the only reliable signal.

Respond in EXACTLY the following format, starting with the line
"# History Summary":

# History Summary

### REASONING
Max 3 telegraphic lines: where the task stands, with provenance tags.

### COMPLETED
One terse line per completed subtask, tagged [O@i]. No values (rule 5).

### PENDING
Remaining steps + known error info, terse. Plans/decision rules tagged
[D@i], stating what they were based on (e.g. "based on agent's reading of
task instruction at T26") -- state provenance only, do NOT add
verification directives or commands.
For EVERY enumeration over a collection, record completeness state:
- "[O@4] web search: first page only, NOT confirmed exhaustive"
- "[O@7] csv rows: read to end of file, COMPLETE"
Never present a partial collection as complete.

### FAILED ATTEMPTS
Only dead ends NOT marked (FAILED) (see rule 6), one terse line each:
tried what -> outcome -> RULED OUT / retry-with-what-change.
Never drop a previously valid entry unless the same action later
succeeded. If none, write "None".

### KEY DATA
ONLY [O@i] values: figures, dates, names, URLs, IDs, lists -- exact
copies, each with source tag. Derived datasets (e.g. "rows to exclude")
go to PENDING as [D@i], not here.

### PINNED CONTENT
One placeholder per line (see pinning rules), e.g.:
<<TURN_3_RESULT>>
If none needed, write "None".

Do NOT repeat the original trajectory. Return ONLY the formatted summary.
\end{lstlisting}

Listing~\ref{lst:officebench_agent_input} illustrates an example of the input to the
specialized agent at decision step $t$.
The same input organization is used across benchmarks.
The system prompt and original task remain fixed throughout an episode, while
the \texttt{HISTORY\_SUMMARY} block is updated only when compression is
triggered.
Its semantic fields are generated by the compressor, whereas the regions
annotated as \texttt{YAMATO-RESTORE} correspond to execution-critical records
restored verbatim from the Append-only Canonical View.
These annotations are included only for exposition; at runtime, the restored
raw content is inserted directly without such markers.
The most recent $K$ interaction turns bypass compression and are retained
verbatim, while the current raw observation and benchmark-specific action or
tool schema are appended for the next decision.
\begin{lstlisting}[
  caption={Input to the specialized agent in OfficeBench at decision step $t$.
  The notation is explanatory and is not included in the runtime prompt.},
  label={lst:officebench_agent_input},
  basicstyle=\ttfamily\scriptsize,
  frame=single,
  breaklines=true,
  columns=fullflexible,
  keepspaces=true,
  showstringspaces=false
]
messages = [

  # [SESSION-FIXED]
  # Instantiated once for the current OfficeBench episode.
  {
    "role": "system",
    "content":
      "You are an AI assistant operating an office automation
       environment. Respond with a single JSON action per step.
       User: <username>
       Date and time: <date, weekday, time>
       Available applications: <application descriptions>
       Workspace: <testbed path>
       ..."
  },

  # [TASK-FIXED + REFRESH-r]
  # The original task remains unchanged. The memory block is replaced
  # whenever compression refresh r completes.
  {
    "role": "user",
    "content":
      "Task: <original OfficeBench instruction>
       Available apps: <initial application list>
       ...

       <HISTORY_SUMMARY>
       [FIXED INTERPRETATION TEMPLATE]
       [O@n] = observation from tool result at turn n;
       [D@n] = unverified agent derivation at turn n;
       [UNSOURCED] = no source recorded.
       This memory records earlier interactions but does not override
       the original task.

       [REFRESH-r: COMPRESSOR-GENERATED SEMANTIC STATE]
       # History Summary

       ### REASONING
       <current progress and decision context>

       ### COMPLETED
       <completed subtasks>

       ### PENDING
       <remaining operations and constraints>

       ### FAILED ATTEMPTS
       <persistent failure ledger maintained by YAMATO>

       ### KEY DATA
       <values with turn-level provenance>

       ### PINNED FILES
       [YAMATO-RESTORE RESULT(i)]
       <exact tool result copied from canonical turn i>

       ### PINNED CONTENT (WORK PRODUCT)
       [YAMATO-RESTORE CURRENT-ARTIFACT]
       <latest write action and subsequent edits, in order>

       </HISTORY_SUMMARY>"
  },

  # [TURN-DYNAMIC: RECENT RAW WINDOW]
  # The latest K turns bypass compression and retain their native form.
  {
    "role": "assistant",
    "content":
      "<think>...</think>
       <action>
       {\"app\": \"excel\",
        \"action\": \"set_cell\",
        \"sheet\": \"<sheet>\",
        \"cell\": \"<cell>\",
        \"value\": \"<value>\"}
       </action>"
  },

  # [STEP-t: CURRENT RAW OBSERVATION]
  {
    "role": "user",
    "content":
      "<verbatim result of the previous OfficeBench action>

       ## Current app: <active application>
       ## Instruction:
       <actions and JSON schemas available in the active app>
       <switch-app and finish-task instructions>
       ## Command:"
  }
]
\end{lstlisting}

\subsubsection{Baseline Implementations}
\label{app:baseline_implementation}

We compare DMC-5 against four representative context-management settings.
Unless otherwise specified, all methods use the same fine-tuned Qwen3.5-27B
agent, the same 50-step interaction cap, and the same cost-accounting rules.
For methods that require an auxiliary compression model, we use
DeepSeek-V4-flash unless stated otherwise.
Method-specific trigger and update policies are preserved by design.

\paragraph{No Compression.}
The agent receives the complete interaction trajectory at every step without
any history reduction or auxiliary compressor.
This setting serves as the full-history reference for both task performance and
end-to-end cost.

\paragraph{FIFO.}
FIFO retains only the most recent five interaction turns and discards all
earlier turns~\citep{yang2024swe}.
It introduces no auxiliary model call and therefore represents a simple
window-based context reduction baseline.
We use the same five-turn window as in prior long-horizon agent compression
evaluations.

\paragraph{AgentDiet.}
AgentDiet~\citep{xiao2026agentdiet} performs inference-time trajectory reduction by revisiting an earlier interaction after a fixed lag and replacing it with a shorter version when redundant, expired, or otherwise unnecessary
content can be removed.
The reduction is produced by a lightweight reflection model conditioned on the target interaction and its surrounding trajectory context.
Due to no official implementation was available, we reproduced the method following the algorithm and prompting strategy described in the paper.
The hyperparameters used are the same as those for DMC-5 or follow the default settings from the original paper.

\paragraph{ACON.}
ACON~\citep{kang2025acon} uses generative history compression in which the current compressed memory is recursively updated with newly accumulated interactions.
We use the official implementation, with only the adaptations required to interface with our agent/compressor endpoints and evaluation configuration.
The agent model, compressor model, maximum interaction steps, and cost accounting are the same as those for DMC-5.

\subsection{Extended Experiments}
\label{app:extended_experiments}

\subsubsection{Run-to-run stability}
\label{sec:stability}

\begin{table}[h]
\centering
\caption{Run-to-run stability across three independent evaluations.
Values are reported as mean $\pm$ standard deviation.}
\label{tab:run_stability}
\small
\begin{tabular}{llcc}
\toprule
Benchmark & Method & Task Quality $\uparrow$ & Cost $\downarrow$ \\
\midrule
AppWorld
& No Compression & $77.0 \pm 2.4$ & $13.71 \pm 0.58$ \\
& DMC-5 & $\mathbf{78.2 \pm 0.9}$ & $\mathbf{10.31 \pm 0.35}$ \\
\midrule
OfficeBench
& No Compression & $87.0 \pm 1.9$ & $5.25 \pm 0.30$ \\
& DMC-5 & $\mathbf{88.0 \pm 1.0}$ & $\mathbf{4.49 \pm 0.14}$ \\
\midrule
8-objective QA
& No Compression & EM $39.1 \pm 1.8$  / F1 $47.6 \pm 3.1$ 
& $35.40 \pm 1.45$ \\
& DMC-5 & EM $\mathbf{44.9 \pm 0.3}$  / F1$\mathbf{54.0 \pm 1.0}$ 
& $\mathbf{25.50 \pm 0.53}$ \\
\bottomrule
\end{tabular}
\end{table}

All main results are averaged over three independent runs.
To quantify stochastic variation, we report the mean and standard deviation of task quality and end-to-end cost across runs.
As shown in Table~\ref{tab:run_stability}, DMC-5 consistently improves average task quality while reducing cost on all three benchmarks.
Moreover, its run-to-run standard deviation is consistently smaller than that of No Compression, indicating improved empirical stability, showing the effectiveness of DMC-5's design that combines full-history access with a compact, integrity-preserving Decision View to reduce sensitivity to irrelevant or corrupted context.

\subsubsection{Hyperparameter Sensitivity}
\label{sec:hyperparameter}

DMC-5 has three main context-control hyperparameters:
the recent raw-turn window $K$, compression interval $\Delta$, and token threshold $T$.
Our default configuration uses $K=1$, $N=3$, and $T=6000$.
We vary one hyperparameter at a time while keeping the other two fixed.

\begin{table}[t]
\centering
\caption{Hyperparameter sensitivity on AppWorld.
We vary one hyperparameter at a time while keeping the others fixed.
Gray rows denote the default configuration used in the main experiments.}
\label{tab:hyperparameter_sensitivity}
\small

\begin{subtable}[t]{0.31\linewidth}
\centering
\caption{Threshold $T$ ($K{=}1$, $N{=}3$)}
\begin{tabular}{ccc}
\toprule
$T$ & Acc. $\uparrow$ & Cost $\downarrow$ \\
\midrule
3000  & 72.4 & 15.28 \\
\rowcolor{gray!20}
6000  & 78.2 & 10.31 \\
8000  & 78.0 & 10.96 \\
10000 & 77.0 & 11.58 \\
\bottomrule
\end{tabular}
\end{subtable}
\hfill
\begin{subtable}[t]{0.31\linewidth}
\centering
\caption{Recent $K$ raw($T{=}6000$, $N{=}3$)}
\begin{tabular}{ccc}
\toprule
$K$ & Acc. $\uparrow$ & Cost $\downarrow$ \\
\midrule
\rowcolor{gray!20}
1 & 78.2 & 10.31 \\
2 & 77.8 & 10.85 \\
3 & 78.0 & 11.23 \\
\bottomrule
\end{tabular}
\end{subtable}
\hfill
\begin{subtable}[t]{0.31\linewidth}
\centering
\caption{Interval $\Delta$ ($T{=}6000$, $K{=}1$)}
\begin{tabular}{ccc}
\toprule
$N$ & Acc. $\uparrow$ & Cost $\downarrow$ \\
\midrule
1 & 70.6 & 19.82 \\
\rowcolor{gray!20}
3 & 78.2 & 10.31 \\
5 & 76.9 & 10.68 \\
\bottomrule
\end{tabular}
\end{subtable}

\end{table}

The results show that DMC-5 is reasonably stable around the default configuration.
For the compression threshold, an overly small $T=3000$ degrades both accuracy and cost efficiency, while increasing $T$ beyond 6000 brings no further quality gain and slightly increases cost.
Varying the recent raw-turn window has only a minor effect on accuracy, but larger $K$ consistently increases cost by retaining more uncompressed context.
The compression interval exhibits the clearest trade-off: compressing every turn ($N=1$) substantially increases cost and also hurts accuracy, whereas a larger interval ($N=5$) slightly reduces task quality without further cost benefit.
Overall, the default setting $T=6000$, $K=1$, and $N=3$ provides the best quality--cost balance among the evaluated configurations.

\subsubsection{YAMATO Integrity}
\label{sec:yamato_appendix}

\paragraph{Integrity Metrics.}
We evaluate whether compression preserves execution-critical historical
information using the same integrity metrics as in the main paper.
Execution-critical information refers to exact historical values that may
affect subsequent tool actions, including email addresses, monetary values,
paths or filenames, UUID/hexadecimal identifiers, access tokens, and URLs.
Table~\ref{tab:integrity_metric_def} summarizes the metric definitions.

\begin{table}[h]
\centering
\caption{Definitions of execution-critical information and integrity metrics.}
\label{tab:integrity_metric_def}
\small
\setlength{\tabcolsep}{5pt}
\begin{tabular}{p{0.29\linewidth}p{0.61\linewidth}}
\toprule
Metric & Definition \\
\midrule

Execution-critical information
& Historical values that may affect subsequent tool actions, including
email addresses, monetary values, paths/filenames, UUID/hex IDs,
access tokens, and URLs. \\

Compression with Errors
& Fraction of compression events that contain at least one corrupted
execution-critical item. \\

Corrupted Execution-Critical Items
& Fraction of all execution-critical items in the compressed state whose
values are inconsistent with the corresponding historical records. \\

Recall of Execution-Critical Items
& Among execution-critical items that are both relevant to the current
action and correct in the historical trajectory, the fraction that are
present in the Decision View when the action is generated. \\

\bottomrule
\end{tabular}
\end{table}

Execution-critical items are extracted automatically using deterministic
pattern-based rules over the six value types defined above.
For corruption, each retained item is compared with its corresponding historical
record, and ambiguous matches are marked as \emph{uncertain} and excluded from
the error statistics.
For recall, an item is considered action-relevant when a later action reuses
the exact historical value or re-invokes the operation that originally produced
it across at least one compression boundary.
The recall denominator therefore contains action-relevant execution-critical
items that are correct in the historical trajectory, while the numerator further
requires that the item remains present in the Decision View when the action is
generated.
\begin{equation}
\mathrm{Recall}
=
\frac{
\#\{\text{action-relevant, correct critical items present in Decision View}\}
}{
\#\{\text{action-relevant, correct critical items}\}
}.
\end{equation}

\begin{equation}
\mathrm{Corruption\ Rate}
=
\frac{
\#\{\text{incorrect critical items}\}
}{
\#\{\text{critical items}\}
}.
\end{equation}

\begin{table}[h]
\centering
\caption{
Detailed integrity analysis on AppWorld.
We compare recursive and append-only history organization with and without
YAMATO. Uncertain cases are excluded from the corruption rate.
}
\label{tab:yamato_integrity_appendix}
\small
\setlength{\tabcolsep}{5pt}
\begin{tabular}{lrrrr}
\toprule
Metric
& ACON
& \makecell{ACON\\w/ YAMATO}
& \makecell{DMC-5\\w/o YAMATO}
& \makecell{DMC-5\\w/ YAMATO} \\
\midrule

Compression events
& 711
& 894
& 2,581
& 2,460 \\

Compression with Errors
& 30/711 (4.2\%)
& 9/894 (1.0\%)
& 77/2,581 (3.0\%)
& \textbf{1/2,460 (0.04\%)} \\

Execution-Critical
Items
& 3,893
& 4,607
& 24,771
& 22,551 \\

Corrupted Execution-Critical
Items
& 48
& 16
& 82
& \textbf{3} \\

Corruption rate
& 1.23\%
& 0.35\%
& 0.33\%
& \textbf{0.013\%} \\

Recall of Execution-Critical
Items
& 63.3\% {\scriptsize(279/441)}
& 87.1\% {\scriptsize(608/698)}
& 77.7\% {\scriptsize(839/1,079)}
& \textbf{98.3\%} {\scriptsize(1,008/1,025)} \\

\bottomrule
\end{tabular}
\end{table}
\paragraph{Integrity Results with and without YAMATO.}
As shown in Table~\ref{tab:yamato_integrity_appendix}, YAMATO consistently improves integrity under both history organizations.
For recursive ACON, it reduces the critical-state error rate from 1.23\% to 0.35\% and raises recall from 63.3\% to 87.1\%.
For append-only DMC-5, the corresponding error rate drops from 0.33\% to 0.013\%, while recall increases from 77.7\% to 98.3\%.
These results indicate that append-only history organization and deterministic state restoration are complementary. 
The former avoids recursive sematic drift, whereas YAMATO protects exact execution-critical values from lossy transcription.

It worth note that the denominators differ across methods because they are defined over action-relevant critical items encountered in each method's own trajectory, which varies with the generated action sequence and interaction length.

\begin{table}[h]
\centering
\caption{Validation of pointer restoration and provenance consistency on AppWorld.}
\label{tab_pointer_provenance}
\small
\begin{tabular}{lr}
\toprule
Metric & Result \\
\midrule
\multicolumn{2}{l}{\textit{Pointer restoration}} \\
Pinned references & 5,181 \\
Valid references & 5,181 / 5,181 (100\%) \\
Successfully resolved & 5,181 / 5,181 (100\%) \\
Byte-exact restoration & 5,169 / 5,181 (99.77\%) \\
Non-byte-exact restoration & 12 / 5,181 (0.23\%) \\
Non-byte-exact cases marked \texttt{[UNSOURCED]} & 12 / 12 (100\%) \\
Silent non-byte-exact restoration & 0 \\
\bottomrule
\end{tabular}
\end{table}
\paragraph{Pointer Restoration Fidelity.}
We first verify whether execution-critical information referenced by pinned
pointers is restored faithfully from the Append-only Canonical View.
Across $168\times3$ AppWorld runs, YAMATO emits 5,181 pinned references.
All references are valid and successfully resolved to the corresponding raw
records.
All 5,181 pinned references are successfully resolved to their original
canonical records, confirming exact deterministic restoration.
Of these, 5,169 remain byte-identical in the final Decision View
(99.77\%).
The remaining 12 cases are also restored correctly, but are subsequently
augmented with an explicit \texttt{[UNSOURCED]} marker by the provenance guard
because the associated claims lack a recognized historical source.
Therefore, these byte-level differences reflect deliberate post-restoration
annotation rather than any corruption of the restored raw content.

\begin{table}[h]
\centering
\caption{Provenance consistency analysis on AppWorld.}
\label{tab:provenance_consistency}
\small
\setlength{\tabcolsep}{5pt}
\begin{tabular}{lrr}
\toprule
Metric & Count & Rate \\
\midrule

Audited claim lines
& 49,700
& -- \\

Claims with provenance tags
& 38,325
& 77.1\% \\

Correctly grounded tagged claims
& 37,991 / 38,325
& 99.13\% \\
\midrule

Correctly downgraded \texttt{[UNSOURCED]} items
& 12 / 12
& 100\% \\

Incorrect \texttt{[O@i]} assignments
& 1 / 35,800
& 0.003\% \\

Agent derivations incorrectly upgraded to \texttt{[O@i]}
& 1 / 2,525
& 0.040\% \\
\bottomrule
\end{tabular}
\end{table}
\paragraph{Provenance Consistency.}
We audit 49,700 claim lines in total.
Among them, 38,325 factual claims carry explicit provenance tags, while the remaining 11,375 lines are mainly structural continuations in multi-line fields rather than standalone factual claims.
Of the tagged claims, 37,991 are supported by the referenced tool output, action, or task instruction, yielding 99.13\% provenance consistency.
The remaining 334 mismatches are dominated by formatting artifacts, naming normalization, and imprecise turn indices rather than unsupported factual content.

We then inspect the three provenance failure modes most relevant to information
integrity.
First, all 12 items that lack a valid raw source are correctly downgraded to
\texttt{[UNSOURCED]}, showing that unsupported information is not silently
treated as grounded state.
Second, among 35,800 claims labeled with observation provenance
\texttt{[O@i]}, only one is incorrectly assigned to a non-existent
turn-level observation, caused by context truncation.
Third, among 2,525 agent-derived claims, only one is incorrectly upgraded from
a derivation to observation provenance.
These results show that YAMATO not only preserves pinned values, but also
maintains a reliable distinction between observed, derived, and unsupported
information.

\begin{table}[h]
\centering
\caption{Detailed analysis of repeated failed actions on AppWorld.}
\label{tab:failure_repetition_appendix}
\small
\setlength{\tabcolsep}{5pt}
\begin{tabular}{lccc}
\toprule
Metric & No Compression & ACON & DMC-5 \\
\midrule
Runs with Repeat $\geq 1$
& 31.3\%
&  \textbf{28.3\%}
& 32.3\% \\

Runs with Same-Action Repeat $\geq 3$
& 4.0\%
& 6.2\%
& \textbf{3.2\%} \\

Runs with Same-Action Repeat $\geq 5$
& 1.2\%
& 1.6\%
& \textbf{0.2\%} \\

Maximum Same-Action Repeats
& 9
& 39
& \textbf{6} \\
\bottomrule
\end{tabular}
\end{table}
\paragraph{Repeated Failed-Action Analysis.}
We further analyze how compression affects repeated actions after explicit execution failures.
Repeated failed actions are detected by normalizing action strings and counting subsequent actions that are identical or have a similarity of at least 0.9 after an explicit execution error.
As shown in Table~\ref{tab:failure_repetition_appendix}, DMC-5 does not reduce the occurrence of ordinary retries, with the rate of runs containing at least one repeated failed action remaining comparable to the baselines.
However, it substantially suppresses more severe repetition.
The fraction of runs with the same failed action repeated at least five times drops to 0.2\%, compared with 1.2\% for No Compression and 1.6\% for ACON.
The same trend appears in the tail, where DMC-5 reduces the maximum same-action repetition to 6, while ACON exhibits an extreme loop with 39 repeats of the same failed action.
These results suggest that YAMATO mainly limits the escalation of failed actions into persistent repetition loops rather than eliminating ordinary retry behavior.

\begin{figure}[h]
    \centering
    \includegraphics[width=0.8\linewidth]{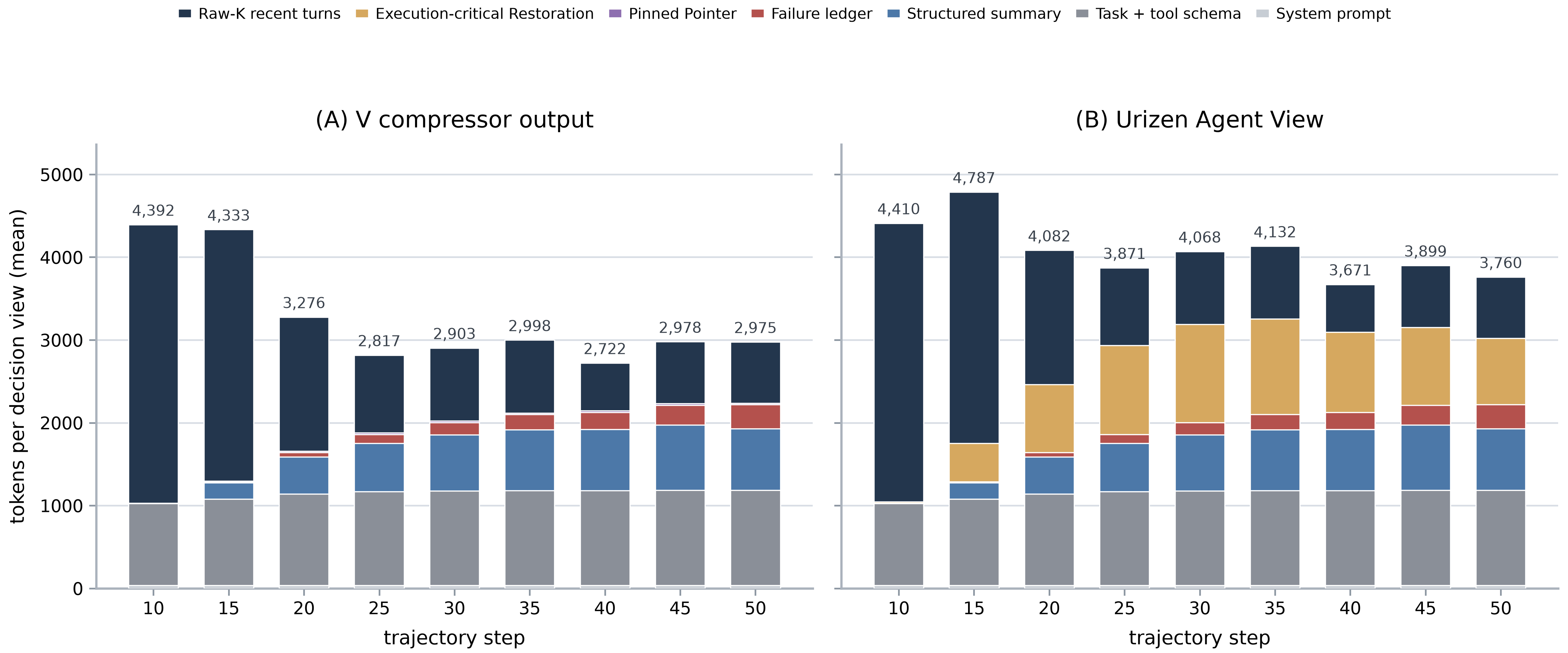}
    \caption{
    Composition of the compressed context over trajectory steps on AppWorld.
    (A) Token composition of V compressor output.
    (B) Token composition of Urizen Decision View after YAMATO restores
    execution-critical records from pinned pointers.
    }
    \label{fig:decision_view_composition}
\end{figure}
\paragraph{Decision View Composition.}
Fig.~\ref{fig:decision_view_composition} decomposes both the compressor output
and the final Urizen Decision View along the trajectory.
Overall, neither view grows with the full interaction history.
After compression is triggered, their total sizes remain within a relatively
stable range despite the continuously growing canonical trajectory, showing
that DMC-5 converts an unbounded interaction history into a bounded working
context for both models.

For the compressor output in Fig.~\ref{fig:decision_view_composition}(A),
the structured summary remains compact as the trajectory grows, while the
failure ledger stays bounded.
The pinned-pointer component is consistently small because execution-critical
raw records are referenced rather than regenerated by the compressor.
This keeps compressor output nearly stable even when more historical state
becomes available in the canonical trajectory.

Fig.~\ref{fig:decision_view_composition}(B) further shows how YAMATO adapts the
agent-facing context over time.
The amount of restored execution-critical information first increases as the
agent accumulates intermediate states and dependencies, but decreases again in
later stages as completed subgoals no longer require the same historical
records.
Thus, DMC-5 does not permanently retain every previously critical item.
Instead, it restores the execution-critical state that remains relevant to the
current decision, while completed or no-longer-needed information is absorbed
into the compact summary.
Together with the bounded failure ledger and recent raw-turn window, this
dynamic restoration keeps the Decision View compact without discarding the
exact state required for subsequent tool execution.

\paragraph{YAMATO Overhead.}
YAMATO consistently improves task performance across both compression architectures, with only modest additional cost under recursive compression and a substantial cost reduction when combined with append-only DMC-5.
As shown in Table~\ref{tab:yamato_main}, on recursive ACON, it improves accuracy from 67.3\% to 72.0\% while slightly increasing total cost from 11.57 to 12.11, although cost per success still decreases from 0.1024 to 0.1001.
On append-only DMC-5, YAMATO improves success from 69.1\% to 78.2\% and simultaneously reduces total cost from 13.31 to 10.31 and cost per success from 0.1147 to 0.0781.
This indicates that integrity preservation is most cost-effective when combined with append-only history organization, where compact pointers and selective restoration avoid repeatedly regenerating or carrying the same historical state.

% baseline：13.71

% 1）仅改变压缩器输入价格（命中折扣固定为0.2）
% P_miss	P_hit	total 
% 0.1	0.02		8.78	
% 0.2	0.04		9.16	
% 0.3	0.06		9.54	
% 0.4	0.08		9.93	
% 0.5	0.10		10.31	
% 0.6	0.12		10.69	
% 0.7	0.14		11.08	
% 0.8	0.16		11.46	
% 0.9	0.18		11.84	
% 1.0	0.20		12.23	

% 2）仅改变压缩器命中折扣（输入未命中价格为0.5）

% discount d	P_hit	total 
% 0.00	0.000		9.46	
% 0.02	0.010		9.54	
% 0.05	0.025		9.67	
% 0.10	0.050		9.88	
% 0.15	0.075		10.10	
% 0.20	0.100		10.31	
% 0.25	0.125		10.52	
% 0.30	0.150		10.74	
% 0.50	0.250		11.59	

% 3）仅改变压缩器输出价格
% P_out		total 
% 0.50		9.98	
% 0.75		10.16	
% 1.00		10.31	
% 1.25		10.46	
% 1.50		10.61	
% 1.75		10.76	
% 2.00		10.91	
% 2.25		11.06	
% 2.50		11.21	

\subsubsection{Performance across Task Difficulty}
\label{sec:difficulty_analysis}

\begin{table*}[h]
\centering
\caption{
AppWorld results stratified by task difficulty.
Agent Input denotes the average cumulative agent input per task, and Input
Slope measures the average growth of agent input with each interaction step.
}
\label{tab:appworld_difficulty}
\small
\setlength{\tabcolsep}{5pt}
\renewcommand{\arraystretch}{1.08}

\begin{tabular}{llccc}
\toprule
Metric & Method & Easy & Medium & Hard \\
\midrule

\multirow{3}{*}{\centering Success}
& No Compression
& 91.2\% & 87.5\% & 57.1\% \\

& \cellcolor{gray!12}DMC-5 w/ DeepSeek-V4-flash
&\cellcolor{gray!12} 91.2\% &\cellcolor{gray!12} 83.3\% &\cellcolor{gray!12} \textbf{63.5\%} \\

& DMC-5 w/ Qwen3.7-flash
& 91.2\% & 70.8\% & 58.7\% \\

\midrule

\multirow{3}{*}{\centering Steps}
& No Compression
& 19.9 & 29.1 & 38.5 \\

&\cellcolor{gray!12} DMC-5 w/ DeepSeek-V4-flash
&\cellcolor{gray!12} 19.7 &\cellcolor{gray!12} 31.3 &\cellcolor{gray!12} 39.1 \\

& DMC-5 w/ Qwen3.7-flash
& 19.9 & 31.3 & 39.4 \\

\midrule

\multirow{3}{*}{\centering Agent Input}
& No Compression
& 133K & 233K & 393K \\

&\cellcolor{gray!12} DMC-5 w/ DeepSeek-V4-flash
&\cellcolor{gray!12} \textbf{85K} & \cellcolor{gray!12}\textbf{138K} & \cellcolor{gray!12}\textbf{187K} \\

& DMC-5 w/ Qwen3.7-flash
& 88K & 142K & 194K \\

\multirow{3}{*}{\centering Input Slope}
& No Compression
& 500 & 461 & 464 \\

& \cellcolor{gray!12}DMC-5 w/ DeepSeek-V4-flash
& \cellcolor{gray!12}\textbf{232} & \cellcolor{gray!12}\textbf{112} & \cellcolor{gray!12}\textbf{98} \\

& DMC-5 w/ Qwen3.7-flash
& 239 & 126 & 118 \\

\midrule

\multirow{3}{*}{\centering Cost / Step}
& No Compression
& 0.00216 & 0.00256 & 0.00324 \\

& \cellcolor{gray!12}DMC-5 w/ DeepSeek-V4-flash
& \cellcolor{gray!12}0.00174 & \cellcolor{gray!12}0.00199 & \cellcolor{gray!12}0.00223 \\

& DMC-5 w/ Qwen3.7-flash
& \textbf{0.00165} & \textbf{0.00180} & \textbf{0.00203} \\

\multirow{3}{*}{\centering Total Cost}
& No Compression
& 2.23 & 3.57 & 7.91 \\

& \cellcolor{gray!12}DMC-5 w/ DeepSeek-V4-flash
& \cellcolor{gray!12}1.82 & \cellcolor{gray!12}2.99 & \cellcolor{gray!12}5.50 \\

& DMC-5 w/ Qwen3.7-flash
& \textbf{1.71} & \textbf{2.70} & \textbf{5.09} \\

\multirow{3}{*}{\centering Cost / Success}
& No Compression
& 0.0429 & 0.0850 & 0.2197 \\

& \cellcolor{gray!12} DMC-5 w/ DeepSeek-V4-flash
&\cellcolor{gray!12} 0.0350 &\cellcolor{gray!12} \textbf{0.0748} &\cellcolor{gray!12} \textbf{0.1375} \\

& DMC-5 w/ Qwen3.7-flash
& \textbf{0.0328} & 0.0794 & 0.1376 \\

\bottomrule
\end{tabular}
\end{table*}

Table~\ref{tab:appworld_difficulty} analyzes DMC-5 across the official
AppWorld difficulty levels.
The efficiency gain becomes more pronounced as task difficulty increases.
With DeepSeek-V4-flash, cumulative agent input is reduced from 133K to 85K on easy
tasks, from 233K to 138K on medium tasks, and from 393K to 187K on hard
tasks.
The corresponding input-growth slopes decrease from 500/461/464 to
232/112/98, showing that DMC-5 increasingly suppresses context growth on
longer and more complex trajectories.

These savings do not come from shortening the interaction.
The number of agent steps remains similar to No Compression across all three
difficulty levels, while cost per step consistently decreases.
Consequently, total cost falls from 2.23/3.57/7.91 to
1.82/2.99/5.50 with DeepSeek-V4-flash.
The reduction is largest on hard tasks, where more accumulated history
provides greater opportunity for compression and prefix reuse.

Task quality exhibits a complementary trend.
DMC-5 preserves the 91.2\% success rate on easy tasks and incurs a moderate
decrease on medium tasks, while improving hard-task success from 57.1\% to
63.5\% with DeepSeek-V4-flash.
Replacing DeepSeek-V4-flash with the cheaper Qwen3.7-flash further reduces cost across
all difficulty levels, but causes a larger quality drop on medium tasks.
This indicates that the efficiency benefit of the dual-view architecture is
consistent across compressors, whereas preserving task quality, particularly
on more demanding trajectories, depends on the compressor's ability to
construct an accurate Decision View.

\subsubsection{Sensitivity to Compressor Pricing}
\label{sec:pricing_sensitivity}

\begin{figure*}[h]
\centering
\includegraphics[width=\textwidth]{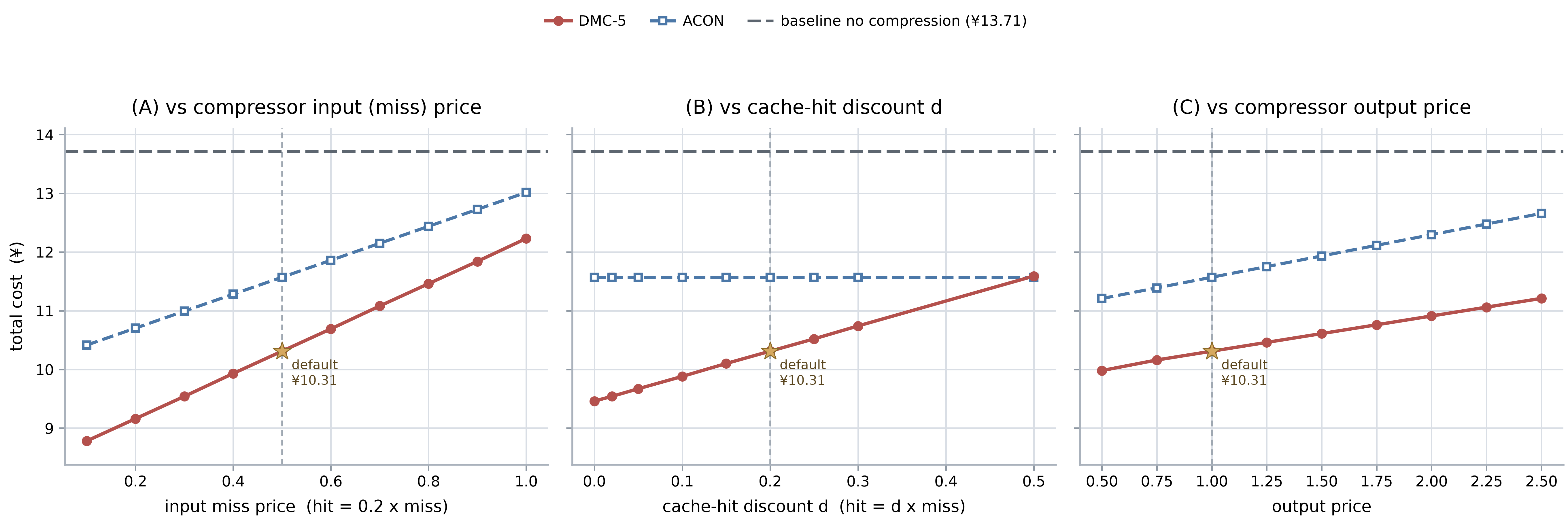}
\caption{
Pricing sensitivity on AppWorld using fixed trajectories.
We recompute billed cost while varying one compressor pricing coefficient at a time and keeping all observed token counts and agent behaviors unchanged.
(A) varies the compressor uncached-input price while keeping the cached-input price at $0.2\times$ the uncached-input price.
(B) varies the cached-input price ratio $d$, where $p^{\mathrm{hit}}=d\,p^{\mathrm{miss}}$.
(C) varies the compressor output price.
Vertical dashed lines denote the default configuration, and the horizontal dashed line denotes the No Compression cost.
}
\label{fig:pricing_sensitivity}
\end{figure*}

We further examine whether the cost advantage of DMC-5 depends on a particular
compressor pricing configuration.
Fig.~\ref{fig:pricing_sensitivity} recomputes the end-to-end cost on fixed
AppWorld trajectories while varying the compressor input price, cached-input
discount, and output price.
The agent trajectories and token counts are kept unchanged, so the analysis
isolates pricing effects without introducing behavioral variation.

Fig~\ref{fig:pricing_sensitivity}(A) examines the compressor uncached-input price.
As the input price increases, the costs of both compression methods increase,
but DMC-5 remains consistently cheaper than ACON and No Compression throughout
the evaluated range.
Notably, DMC-5 still provides a net saving even when the compressor uncached-input
price exceeds that of the specialized agent.
This is because most historical compressor input is served through the cheaper
cached-input path, so a relatively expensive compressor does not necessarily
translate into a high end-to-end cost.

Fig~\ref{fig:pricing_sensitivity}(B) isolates the effect of the cached-input price ratio.
ACON is almost unaffected because its recursively rewritten summaries provide
little reusable prefix, whereas DMC-5 becomes progressively cheaper as the
cached-input price decreases.
Even when cached input is priced at 50\% of uncached input, DMC-5 remains below
the No Compression baseline and approximately matches ACON.
The gain therefore does not require an extreme cache discount, while stronger
discounts directly translate the high prefix-reuse rate of DMC-5 into larger
cost savings.

Fig~\ref{fig:pricing_sensitivity}(C) varies the compressor output price.
DMC-5 remains cheaper than both ACON and No Compression over the entire range,
including settings where compressor output is priced comparably to or even
higher than specialized-agent output.
This robustness follows from the compact compressor output enabled by structured
summaries and pinned pointers, which limits the amount of expensive generated
content.

Together, these results show that DMC-5 is not tied to a narrow pricing
configuration.
Its cost advantage persists across substantially different input prices,
cache discounts, and output prices, and is consistently stronger than recursive
compression.

\subsubsection{Wall-Clock Latency}
\label{sec:latency}

DMC-5 optimizes billed inference cost rather than minimum response latency.
Because each compression event introduces an additional model invocation,
the monetary savings are accompanied by increased wall-clock time.
Table~\ref{tab:wallclock_appendix} reports the measured average end-to-end
latency per task.

\begin{table}[h]
\centering
\caption{Average wall-clock latency per task (seconds).}
\label{tab:wallclock_appendix}
\small
\begin{tabular}{lccc}
\toprule
Method & AppWorld & OfficeBench & 8-objective QA \\
\midrule
No Compression & 69  & 38 & 122 \\
ACON           & 78  & 42 & 149 \\
DMC-5          & 94 & 44 & 186 \\
\bottomrule
\end{tabular}
\end{table}

The overhead is smallest on the shorter OfficeBench workload and larger on
longer trajectories that invoke the compressor more frequently.
Thus, DMC-5 exposes a clear cost--latency trade-off: it substantially reduces
billed inference cost, but is not designed to minimize serial response time.
Parallel or asynchronous compression is orthogonal to the proposed context
representation and remains an interesting direction for deployment-oriented
optimization.

\subsubsection{Representative Trajectories}
\label{sec:cases}

We qualitatively inspect representative AppWorld trajectories to illustrate how
compression errors affect downstream actions.
Table~\ref{tab:case_summary} summarizes three cases in which recursive
compression loses or distorts task state while DMC-5 succeeds, together with
one DMC-5 failure case that illustrates the boundary of context preservation.

\begin{table*}[h]
\centering
\caption{Representative AppWorld trajectories.
The first three cases illustrate compression-induced failures avoided by
DMC-5, while the last case shows a failure outside the scope of YAMATO.}
\label{tab:case_summary}
\small
\setlength{\tabcolsep}{5pt}
\begin{tabular}{p{0.12\linewidth}p{0.19\linewidth}p{0.29\linewidth}p{0.30\linewidth}}
\toprule
Task & Outcome & Failure mode & DMC-5 behavior \\
\midrule

\texttt{986aa4e\_1}
& ACON fails at 50 steps;
DMC-5 passes 3/3
& An access token disappears across recursive summaries, forcing re-login and
causing a previously failed login action to be repeated.
& YAMATO preserves the exact token in \textsc{Key Data} with provenance,
avoiding re-login and repeated failure. \\

\texttt{09b0ee6\_2}
& ACON fails at 50 steps;
DMC-5 passes 3/3
& A recursive summary regresses from ``all 83 songs processed'' to
``83 song IDs ready for lookup'', discarding accumulated aggregation state.
& Append-only reconstruction avoids recursive state regression, while the
final aggregate result remains grounded in \textsc{Key Data}. \\

\texttt{634f342\_1}
& ACON fails at 50 steps;
DMC-5 passes 2/3
& API documentation, access tokens, and completed file-processing state are
lost or distorted, causing repeated documentation lookup, re-login, and
re-execution.
& Pinned raw records and grounded state preserve API results, credentials,
and completed progress across compression rounds. \\

\texttt{9016950\_1}
& DMC-5 fails at 24 steps
& All required historical information is preserved, but the agent incorrectly
resolves the scope of ``them'' and performs an unnecessary action.
& The failure originates from task interpretation rather than context loss,
illustrating the boundary of YAMATO. \\

\bottomrule
\end{tabular}
\end{table*}

\textbf{Case 1  Preserving exact execution state.}
Task \texttt{986aa4e\_1} requires the agent to modify a Spotify playlist
according to comments in Todoist, leave a confirmation comment, and mark the
task as completed.
ACON reaches the 50-step interaction limit, whereas DMC-5 succeeds in all
three runs.

The failure begins when recursive compression retains the statement that the
agent is already logged in but drops the exact Spotify access token.
The downstream agent therefore knows that authentication has occurred but no
longer has the credential required for subsequent API calls.

\begin{lstlisting}[basicstyle=\ttfamily\scriptsize,breaklines=true]
ACON summary
- Access token from earlier login is stored in session; no need to re-login.

Agent
[27] According to the history summary, I logged in successfully ...
[28] I need the access_token ... but I don't have the access token
[29] I need to log in to Spotify to get the access token ...
\end{lstlisting}

Re-authentication also causes the agent to repeat a previously encountered
argument error.

\begin{lstlisting}[basicstyle=\ttfamily\scriptsize,breaklines=true]
[14] login(email=...)        -> Unexpected parameter "email"
[15] login(username=...)     -> success
...
[30] login(email=...)        -> the same error appears again
[32] login(username=...)     -> success
\end{lstlisting}

DMC-5 instead preserves the exact credentials and task identifiers in
\textsc{Key Data}, together with their observation provenance.

\begin{lstlisting}[basicstyle=\ttfamily\scriptsize,breaklines=true]
### KEY DATA
- Todoist access_token: eyJ... [O@7]
- Spotify access_token: eyJ... [O@15]
- Project id: 280 [O@10]
- Task id: 6342 [O@11]
- Playlist id: 654 [O@16]
- Songs added: 248, 160, 65, 140, 115 [O@19-23]
- Songs removed: 260, 49 [O@16, O@33-34]
\end{lstlisting}

The agent therefore reuses the original token instead of re-authenticating,
avoids repeating the failed login pattern, and completes the remaining
playlist operations and task update.
This case illustrates how preserving exact execution state prevents a local
compression error from propagating into repeated interactions.

\textbf{Case 2  Preventing recursive state regression.}
Task \texttt{09b0ee6\_2} asks the agent to identify the four artists with the
fewest unique songs across all of the user's Spotify playlists.
The answer requires aggregating metadata from 83 unique songs.
ACON reaches the interaction limit without producing the answer, while
DMC-5 succeeds in all three runs.

Before the fifth compression, the recursive summary correctly records that all
83 songs have already been retrieved and that only aggregation remains.

\begin{lstlisting}[basicstyle=\ttfamily\scriptsize,breaklines=true]
Earlier ACON summary
- Added Marcus Lane ..., Hazel Winter ..., Lucas Grey ...
- All 83 songs fetched. Ready to aggregate artist counts ...
\end{lstlisting}

After another recursive update, however, the same task state regresses to an
earlier stage.

\begin{lstlisting}[basicstyle=\ttfamily\scriptsize,breaklines=true]
Later ACON summary
- Compiled 83 unique song IDs ready for artist lookup.
\end{lstlisting}

The accumulated artist statistics are no longer available, so the agent begins
retrieving the song metadata again and eventually exhausts the 50-step budget.

DMC-5 avoids this failure because each compressed state is reconstructed from
the append-only raw trajectory rather than from the previous summary.
Once the aggregation has been completed, the resulting state is explicitly
grounded and retained.

\begin{lstlisting}[basicstyle=\ttfamily\scriptsize,breaklines=true]
### KEY DATA
- 4 least-song artists (1 song each):
  Grace Harrison, Luna Starlight, Evelyn Rose, Liam Palmer [O@21]

### COMPLETED
- Collected 83 unique song IDs across all playlists [O@19]
- Counted songs per artist (28 artists) [O@20]
- Identified 4 artists with least songs [O@21]

### PENDING
- Submit answer:
  "Grace Harrison, Luna Starlight, Evelyn Rose, Liam Palmer" [D@21]
\end{lstlisting}

The agent subsequently submits the result without repeating the enumeration.
This case demonstrates that append-only reconstruction prevents completed
progress from being recursively rewritten into an earlier task state.

\textbf{Case 3  Preserving tool records and completed progress.}
Task \texttt{634f342\_1} requires the agent to read a file containing 14 old
songs, locate them across Spotify playlists, move them into a new
``Archived Songs'' playlist, and remove them from their original playlists.
ACON reaches the 50-step limit, whereas DMC-5 succeeds in two of three runs.

Under recursive compression, several forms of operational state disappear.
The agent repeatedly reloads API documentation that had already been observed.

\begin{lstlisting}[basicstyle=\ttfamily\scriptsize,breaklines=true]
[13] show_api_descriptions(app_name="spotify")
[17] show_api_descriptions(app_name="spotify")
...
[37] show_api_descriptions(app_name="spotify")
[38] show_api_doc(app_name="spotify", ...)
\end{lstlisting}

The exact access token is also lost, leading the agent to reconstruct or
reacquire it.

\begin{lstlisting}[basicstyle=\ttfamily\scriptsize,breaklines=true]
[35] spotify_token = "BQDh2JjK8vN5mR7pL3xW9..."
     -> unterminated / truncated string
...
[40] The spotify_token variable was lost.
     I need to get it again by logging in.
\end{lstlisting}

More importantly, an operation that has already succeeded is later rewritten
as unfinished.

\begin{lstlisting}[basicstyle=\ttfamily\scriptsize,breaklines=true]
Earlier state
- old_songs.txt successfully read and parsed into 14 songs

Later recursive summary
- Attempted to read old_songs.txt but failed initially ...
- Need to retry reading old_songs.txt with the access token.
\end{lstlisting}

DMC-5 retains exact credentials and completed state in grounded
\textsc{Key Data}, while large tool records can be referenced through pinned
raw results.

\begin{lstlisting}[basicstyle=\ttfamily\scriptsize,breaklines=true]
### KEY DATA
- file_system access_token: eyJ... [O@12]
- spotify access_token: eyJ... [O@16]
- old_songs.txt content: 14 songs listed [O@14]
- Matched song IDs: [29, 43, 74, ..., 311] [O@29]
- Archived Songs playlist ID: 654 [O@34]

### PINNED DOCS
<<TURN_40_RESULT>>
\end{lstlisting}

The agent therefore avoids repeatedly reconstructing already observed tool
state and can continue from the preserved task progress.
This case illustrates why exact raw restoration is particularly useful for
operational records that are difficult to reproduce reliably through natural
language summaries.

\textbf{Case 4  Failure outside the scope of context preservation.}
Task \texttt{9016950\_1} asks the agent to create a Venmo account for each
parent who does not already have one and then notify them by text message.
DMC-5 terminates normally after 24 steps but passes only 6 of 7 checks.

The agent correctly identifies that Chad does not have a Venmo account while
Sabrina already has one, and therefore creates an account only for Chad.
The subsequent failure occurs when the agent incorrectly expands the reference
of ``them'' to both parents.

\begin{lstlisting}[basicstyle=\ttfamily\scriptsize,breaklines=true]
[19] search_users(Chad)      -> no Venmo account
[20] search_users(Sabrina)   -> existing Venmo account
[21] signup(Chad, ...)       -> success

[22] "Now I need to send text messages to both parents."
     send_text_message(Chad, ...)

[23] send_text_message(Sabrina, ...)   <- unnecessary action
[24] complete_task()
\end{lstlisting}

This failure is not caused by missing compressed state.
The contact information, account status, message content, and authentication
credentials remain available throughout the trajectory, and the agent does not
repeat any login or information-retrieval operation.
Instead, the error arises from incorrect referential reasoning over the task
instruction.

The case therefore illustrates a clear boundary of DMC-5.
YAMATO protects historical information from omission or corruption during
compression, but it cannot correct semantic reasoning errors made by the
underlying agent when the required information is already available.

Across the three successful cases, recursive compression fails through
different manifestations of information loss, including the loss of exact
execution-critical values, regression of completed task state, and omission
of previously observed tool records.
DMC-5 mitigates these failures through the Append-only Canonical View,
the compact Decision View, and YAMATO's deterministic restoration of
execution-critical information.
The final case further illustrates the scope of the method.
YAMATO preserves information integrity during lossy compression, but reasoning
failures caused by limitations of the underlying agent model are outside the
scope of DMC-5 and do not directly benefit from these mechanisms.

\subsection{Limitation and Future Work}

Our experiments show that DMC-5 remains effective across different compressor models, task difficulties, and a broad range of pricing configurations, while YAMATO substantially mitigates compression-induced information loss and repeated failures. 
Nevertheless, several broader settings remain beyond the current design scope.

First, DMC-5 is designed primarily to reduce billed inference cost by separating full-history integration from specialized action generation and exploiting prefix reuse in the compressor. 
The current implementation already achieves substantial cost savings, but compression is performed synchronously, so the additional compressor calls can increase wall-clock latency. 
Further reducing this latency overhead remains an open problem beyond the current scope of DMC-5.
Future work could therefore investigate asynchronous or parallel compression, speculative refresh, and overlap between compression and agent execution.

Second, YAMATO effectively protects execution-critical information from corruption or omission during compression. However, when the required information is already available, the underlying agent may still make task-interpretation or reasoning errors. Handling such reasoning errors remains beyond the current scope of DMC-5 and may benefit from stronger planning, verification, or self-correction mechanisms.

Third, DMC-5 currently uses a fixed compression policy with predefined thresholds and a fixed division of roles between the general compressor and specialized agent. 
Our sensitivity experiments indicate that the default policy is robust across the evaluated settings, but different workloads may favor different compression timing, context allocation, or compressor choices. 
Adapting compression behavior to workload characteristics remains an open direction beyond the current scope of DMC-5.
Future work could explore adaptive compression triggers, dynamic model selection, and tighter integration with serving-side cache management, as well as broader evaluation on longer-horizon and more heterogeneous agent workloads.

\subsection{Related Work}
\label{sec:extended_related_work}

Section~\ref{sec:related_work} provides a compact comparison in the main paper.
Here we further distinguish related methods by the state they operate on,
how that state is updated over time, and where computational savings arise.

\subsubsection{Agent Context Compression and Memory}

\paragraph{Prompt and context compression.}
Prompt-compression methods primarily reduce the representation presented to a
downstream model.
LLMLingua scores and removes less informative tokens, while LongLLMLingua
introduces query-aware compression for long inputs
~\citep{jiang2023llmlingua,jiang2023longllmlingua}.
LLMLingua-2 learns a bidirectional token classifier from distilled compression
data, Dodo performs input-dependent contextual compression for decoder-only
models, and RECOMP selects or summarizes retrieved evidence before generation
~\citep{pan2024llmlingua2,qin2024dodo,xu2023recomp}.
These methods mainly optimize the length or information density of a given
model input.

A complementary line compresses text into learned latent representations.
Gisting encodes reusable information into special tokens, AutoCompressors
recursively construct summary vectors, and ICAE maps long inputs into compact
memory slots
~\citep{mu2023gist,chevalier2023autocompressors,ge2024icae}.
Latent representations can achieve high compression density, but generally
require a compatible downstream model.
DMC-5 instead produces a textual Decision View that can be consumed by an
independently deployed specialized agent.
More importantly, DMC-5 focuses on repeated trajectory integration rather than
compressing a single static input.

\paragraph{Persistent agent memory.}
Agent-memory systems focus on organizing historical information for later
access.
MemGPT moves information across memory tiers, HiAgent groups interactions by
subgoal, Generative Agents retrieves experiences and forms higher-level
reflections, and Reflexion stores verbal feedback from previous failures
~\citep{packer2023memgpt,hu2024hiagent,park2023generativeagents,shinn2023reflexion}.
MEMORYLLM maintains an updatable latent memory pool, while RAPTOR organizes
recursive summaries in a retrieval hierarchy
~\citep{wang2024memoryllm,sarthi2024raptor}.
These methods primarily address how historical information is stored,
organized, or retrieved.
DMC-5 addresses a different problem: how a growing interaction trajectory is
repeatedly integrated into the next decision while controlling both
agent-side input and compressor-side cost.

\paragraph{Trajectory-update strategies.}
Recent agent context-management methods can be characterized by their dominant
history-update behavior.
These categories describe update patterns rather than mutually exclusive
method families.

Selective-update methods retain a recent window or selected trajectory
elements while removing or folding less relevant content
~\citep{xiao2026agentdiet,ning2026agentomit,hu2024hiagent,ye2026agentfold,verma2026focus}.
They reduce the amount of history processed at each step, but information
removed from the active state may be unavailable when it becomes relevant
again.

Recursive-update methods maintain a bounded state by combining the previous
compressed memory with newly generated interactions
~\citep{zhou2025mem1,yu2025memagent,wu2025resum}.
This avoids rereading the complete trajectory, but repeatedly rewriting the
same state can accumulate omissions or altered details
~\citep{kang2025acon,shi2025rememr1}.
It also changes the prefix presented to the compressor across rounds, limiting
prefix-cache reuse.

Full-history recompression instead reconstructs a compact representation from
retained raw context, preserving access to earlier information but repeatedly
processing an expanding history
~\citep{li2023selectivecontext,jiang2023longllmlingua,xu2023recomp}.

DMC-5 combines full-history access with a different update protocol.
Its Append-only Canonical View preserves raw trajectory records and a stable
historical prefix for repeated compressor calls, while a separate compact
Decision View is regenerated for the specialized agent.
YAMATO further restores selected execution-critical records directly from the
Canonical View rather than relying on model rewriting.
The design therefore jointly targets history recoverability, compression
integrity, prefix-cache reuse, and agent-side context cost.

\subsubsection{Cache-Aware LLM Inference}

\paragraph{Prefix reuse and serving-side cache organization.}
LLM serving systems reduce repeated computation by organizing reusable KV
states.
PagedAttention manages KV blocks using virtual-memory techniques, SGLang
organizes shared prefixes in a radix tree, Prompt Cache exposes reusable prompt
modules, and FastTree optimizes attention execution for prefix-sharing request
trees
~\citep{kwon2023pagedattention,zheng2024sglang,gim2024promptcache,pan2025fasttree}.
These systems optimize how reusable states are stored or scheduled once shared
prefixes exist.

DMC-5 operates at a different layer.
It does not modify the serving runtime or directly manipulate KV tensors.
Instead, it structures repeated compressor requests so that the historical
portion remains byte-identical across rounds.
The application therefore creates a stable reusable prefix that standard
provider-side prefix caching can exploit.

\paragraph{KV-cache selection and compression.}
Another line reduces the memory or bandwidth footprint of cached attention
states.
H$_2$O retains recent tokens and attention heavy hitters, Scissorhands exploits
persistent token importance, StreamingLLM combines attention sinks with a recent
window, SnapKV selects head-specific KV positions, QUEST loads query-relevant
KV pages, and ThinKV combines reasoning-aware retention with adaptive
quantization and eviction
~\citep{zhang2023h2o,liu2023scissorhands,xiao2024streamingllm,
li2024snapkv,tang2024quest,ramachandran2025thinkv}.

These methods compress or select internal model state.
They can reduce inference memory or attention cost without necessarily reducing
the prompt tokens billed to an API endpoint.
DMC-5 instead reduces the textual context repeatedly processed by the acting
model while organizing compressor input for prefix reuse.
The two optimization layers are therefore complementary.

\paragraph{Cache reuse across requests and workflows.}
CacheGen encodes and transfers KV states across machines, CacheBlend reuses
non-prefix cached chunks with selective recomputation, KVFlow schedules prefix
retention across agent workflows, and LRAgent shares base-model cache states
across LoRA-specialized agents
~\citep{liu2024cachegen,yao2024cacheblend,pan2025kvflow,jeon2026lragent}.
These methods improve how reusable cache states are transferred, scheduled, or
shared across requests.

CAPC is particularly related to DMC-5 because it explicitly studies the
conflict between prompt compression and prefix reuse
~\citep{song2026capc}.
CAPC optimizes compression within an individual prompt so that shortening the
input does not unnecessarily destroy reusable cached prefixes.
DMC-5 addresses the complementary multi-turn problem.
It organizes a growing trajectory across repeated compressor calls so that the
historical prefix remains reusable, while the specialized agent receives a
separate Decision View optimized for action generation.
Thus, cache management and semantic compression remain separate mechanisms,
but DMC-5 connects them through the trajectory-update protocol.

\subsubsection{Multi-Model Collaboration and Agent Efficiency}

\paragraph{Routing and cascades.}
Model routers and cascades reduce inference cost by deciding which model should
handle a request.
FrugalGPT learns cascades across model APIs, RouteLLM learns routing decisions
from preference data, Hybrid LLM predicts query difficulty, and AutoMix starts
with a cheaper model and escalates uncertain requests
~\citep{chen2023frugalgpt,ong2024routellm,ding2024hybridllm,
aggarwal2024automix}.
Their central decision is which model should produce the answer.

DMC-5 does not route requests between interchangeable models.
Both models remain active and have fixed asymmetric roles.
The general compressor performs full-history integration, while the specialized
agent performs downstream action generation.
This functional decomposition also allows each model to receive a context
designed specifically for its role.

\paragraph{Collaborative generation.}
Other multi-model systems collaborate by exchanging generated outputs.
LLM-Blender ranks and fuses candidate responses, Mixture-of-Agents propagates
responses through layered model groups, DUET transfers a compact reasoning
signal between models, and speculative decoding uses a smaller model to draft
tokens that a larger model verifies
~\citep{jiang2023llmblender,wang2025moa,chen2026duet,leviathan2023speculative}.
These methods collaborate over answers, reasoning signals, or token proposals.

DMC-5 instead collaborates over persistent agent state.
The compressor constructs the Decision View from the evolving trajectory, and
the specialized agent acts from that state.
The collaboration therefore persists across interaction turns rather than
forming a one-shot generation pipeline.

\paragraph{Agent evaluation and end-to-end efficiency.}
Long-horizon agent benchmarks expose interaction costs that are largely absent
from single-turn evaluation.
AgentBench spans multiple interactive environments, AgentBoard reports
intermediate progress, WebArena evaluates agents on reproducible websites, and
SWE-bench evaluates repository-level software-engineering workflows
~\citep{liu2024agentbench,ma2024agentboard,zhou2024webarena,
jimenez2024swebench}.
AppWorld and OfficeBench extend this setting to executable API workflows and
office automation
~\citep{trivedi2024appworld,wang2024officebench}.

Final task success alone does not capture the cost of additional model calls,
retries, or longer trajectories.
Agentless demonstrates that simpler pipelines can outperform more elaborate
agents at lower cost, while broader work on agent evaluation argues for
reporting resource consumption together with task quality
~\citep{xia2025agentless,kapoor2024agents}.
DMC-5 follows this perspective by jointly measuring task performance,
agent-side input, compressor cache reuse, and end-to-end billed cost, thereby
capturing both compression savings and interaction overhead caused by lost
state.

\end{document}